\documentclass[nonatbib, preprint,12pt]{elsarticle}

\usepackage{amssymb}

\usepackage{amsmath,amsfonts}
\usepackage{algorithmic}
\usepackage{algorithm}
\usepackage{array}
\usepackage[caption=false,font=normalsize,labelfont=sf,textfont=sf]{subfig}
\usepackage{textcomp}
\usepackage{stfloats}
\usepackage{url}
\usepackage{verbatim}
\usepackage{graphicx}
\usepackage{hyperref} 
\usepackage{color,soul}
\definecolor{red}{rgb}{1.00,0.00,0.00}
\definecolor{blue}{rgb}{0.00,0.00,1.00}
\definecolor{green}{rgb}{0.30, 0.50,0.00}

\newcommand{\cblue}[1] {\textcolor{blue}{#1}}

\usepackage{bm}
\usepackage{booktabs} 
\usepackage{tabularx}  
\usepackage{mwe}
\usepackage{wrapfig}
\usepackage{pifont}
\usepackage[authoryear]{natbib}
\usepackage{xcolor}      
\hypersetup{
    colorlinks=true,
    citecolor=blue,     
    linkcolor=blue,     
    urlcolor=blue       
}
\setcitestyle{authoryear,round}

\journal{Elsevier}

\begin{document}

\begin{frontmatter}



\title{Fast Trajectory Planner with a Reinforcement Learning-based Controller for Robotic Manipulators}


\author[inst1]{Yongliang Wang} 

\author[inst1]{Hamidreza Kasaei\corref{cor1}}
\ead{hamidreza.kasaei@rug.nl}

\affiliation[inst1]{organization={Department of Artificial Intelligence, Bernoulli Institute, Faculty of Science and Engineering, University of Groningen},
    addressline={Nijenborgh 9}, 
    city={Groningen},
    postcode={9747 AG}, 
    country={Netherlands}}

\cortext[cor1]{Corresponding author}

\begin{abstract}
Generating obstacle-free trajectories for robotic manipulators in unstructured and cluttered environments remains a significant challenge. Existing motion planning methods often require additional computational effort to generate the final trajectory by solving kinematic or dynamic equations. This paper highlights the strong potential of model-free reinforcement learning methods over model-based approaches for obstacle-free trajectory planning in joint space. We propose a fast trajectory planning system for manipulators that combines vision-based path planning in task space with reinforcement learning-based obstacle avoidance in joint space. We divide the framework into two key components. The first introduces an innovative vision-based trajectory planner in task space, leveraging the large-scale fast segment anything (FSA) model in conjunction with basis spline (B-spline)-optimized kinodynamic path searching. The second component enhances the proximal policy optimization (PPO) algorithm by integrating action ensembles (AE) and policy feedback (PF), which greatly improve precision and stability in goal-reaching and obstacle avoidance within joint space. These proximal policy optimization (PPO) enhancements increase the algorithm’s adaptability across diverse robotic tasks, ensuring consistent execution of commands from the first component by the manipulator, while also enhancing both obstacle avoidance efficiency and reaching accuracy. The experimental results demonstrated the effectiveness of proximal policy optimization (PPO) enhancements, as well as simulation-to-simulation (Sim-to-Sim) and simulation-to-reality (Sim-to-Real) transfer, in improving model robustness and planner efficiency in complex scenarios. These enhancements allowed the robot to perform obstacle avoidance and real-time trajectory planning in obstructed environments. \href{https://sites.google.com/view/ftp4rm/home}{\cblue{https://sites.google.com/view/ftp4rm/home}}
\end{abstract}

\begin{keyword}
Reinforcement Learning \sep 
Artificial Intelligence Enabled Robotics \sep 
Motion Planning \sep 
Artificial Intelligence Based Methods \sep 
Collision Avoidance
\end{keyword}
\end{frontmatter}



\section{Introduction}
\label{sec:introduction}

Robotic manipulators are widely used in industrial production lines and service robots to assist humans. This integration underscores the need for safe and efficient motion planning to achieve precise manipulations in complex environments \citep{kirner2024impact, nishio2023design}.  The motion planning process consists of four key stages: \citep{wang2024impact}: (1) \textit{\textbf{Task Planning}}, which involves designing a set of high-level goals, for example, $``$\textit{Move the box in front of you to the opposite side of the table.}”; (2) \textit{\textbf{Path Planning}}, the generation of a feasible path from a start to a goal; (3) \textit{\textbf{Trajectory Planning}}, the creation of a time schedule for following the path, given constraints such as position, velocity, and acceleration; and (4) \textit{\textbf{Trajectory Tracking}}, executing the trajectory accurately based on a control system. 

The majority of research has been concentrated on path planning. Graph search and sampling-based methods are two primary approaches. Graph-based algorithms such as A$^*$ and Dijkstra create a graph of the environment to find the most efficient or shortest path from start to goal \citep{merikh2024ergonomically}. Despite their efficiency, these algorithms struggle with computational complexity, memory usage, and adapting to dynamic environments. In contrast, sampling-based methods such as the Probabilistic Roadmap (PRM) \citep{rosmann2015planning, rosmann2017integrated} and the Rapidly-exploring Random Trees (RRT) are capable of navigating collision-free paths in both static and dynamic environments \citep{844730, ji2023rrt, saccuti2023protamp}. Nevertheless, the inherent randomness of the search process in these methods might not always yield the optimal solution, with their effectiveness heavily reliant on the sampling strategy. These approaches often face challenges in constrained environments, and their execution time cannot be guaranteed.

Operating in the task space, i.e., the Cartesian space of the end-effector for manipulators, simplifies motion planning significantly. However, these trajectories only account for obstacle avoidance in the task space. For manipulators, consideration of joint space is also essential. Research on trajectory generation for autonomous vehicles and aerial drones has developed efficient methods that quickly produce high-quality trajectories in cluttered environments. Transferring high-quality trajectory generation from aerial drones to manipulators requires a controller for avoiding obstacles in joint space \citep{zhou2019robust, zhou2020robust, zhou2021raptor, wang2024impact}. A task-space waypoint can be converted to an obstacle-free robot's configuration in joint-space using reinforcement learning (RL) \citep{chen2024trakdis, li2021general, zhang2021sim2real}. Such approaches have been effectively applied to robot manipulators for joint-level control in both simulations and real-world scenarios. Most RL-based research typically limits robot motion to a small portion of the workspace, despite achieving high rewards and task success rates. However, applying these results to real-world environments is challenging due to the larger workspace sizes.

\begin{figure}[!t]
      \centering
      \includegraphics[width=1.0\linewidth]{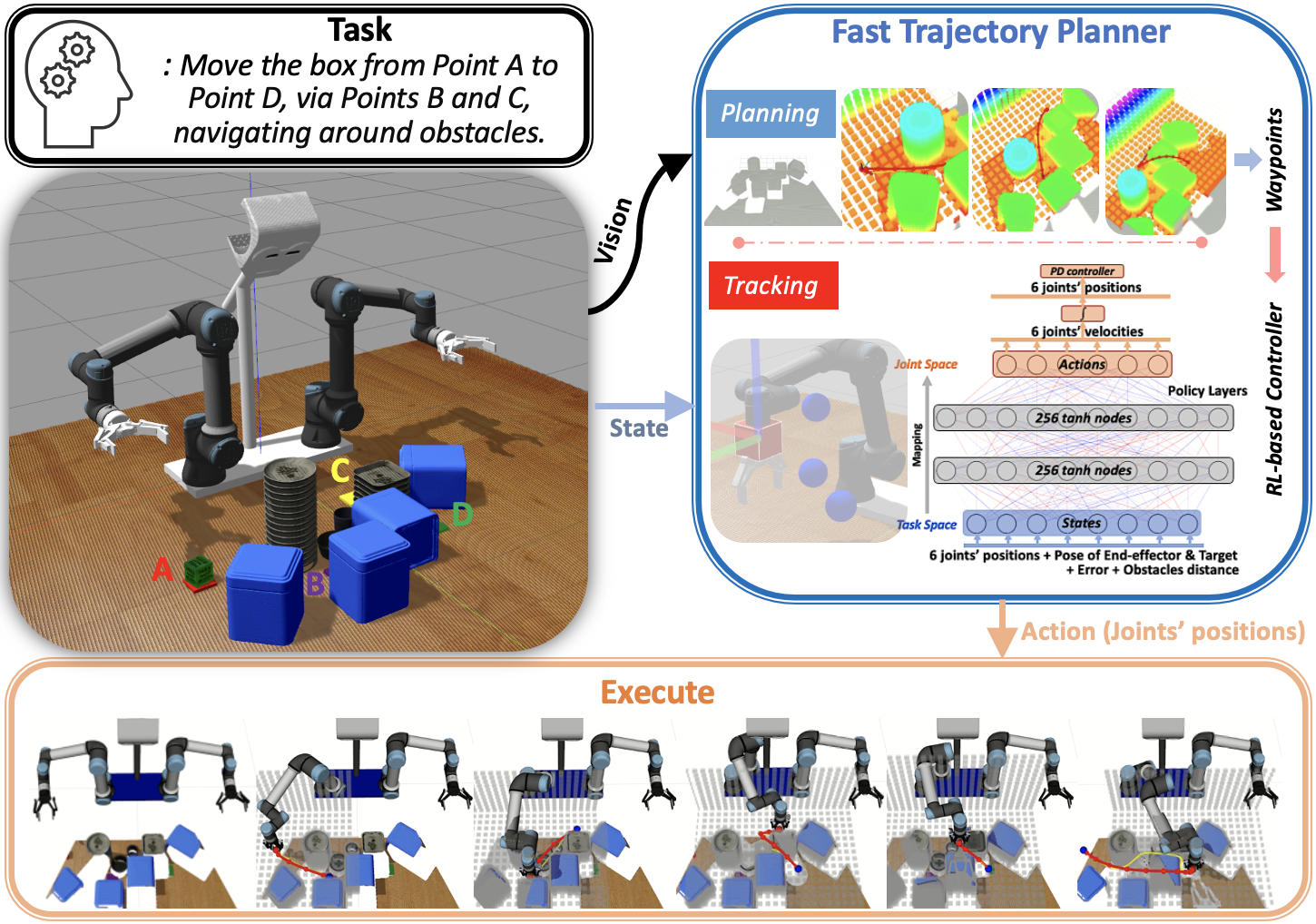}
      \caption{\textbf{System Overview}: The Kinect sensor captures visual data to create an FSA-enhanced 3-dimensional (3D) map of the environment. This map is used with B-spline optimization to generate waypoints for kinodynamic path searching. These waypoints, along with the manipulator's state, inform the RL-based controller that adjusts the positions of the 6 joints of the manipulator.}
      \label{fig:system_overview}
\end{figure}

Although various motion planners exist for manipulators, there remains a lack of a fast, systematic trajectory planning framework tailored specifically for robotic arms. Current planners rarely leverage learning-based approaches to directly map target space configurations to joint space, and often lack the flexibility to be easily integrated and deployed across different manipulator platforms. Furthermore, while RL offers great potential for adaptive control, existing methods are not fully tailored for high-precision reaching tasks in manipulation. In particular, policy-gradient-based algorithms such as PPO have shown promising results in robot control, but still suffer from instability, high action variance, and limited adaptability in dynamic or cluttered environments. These challenges are especially pronounced in tasks requiring precise positioning, rapid re-planning, or coordinated motion under partial observability. Therefore, the central problem addressed in this paper is twofold: (i) How can we design a fast and efficient trajectory planning and tracking framework specifically tailored for manipulators? (ii) How can we leverage learning-based methods to enable efficient mapping from task space to joint space and adapt reinforcement learning algorithms to better suit robotic tasks, particularly high-precision reaching?

In this paper, a comprehensive framework is presented to generate safe and efficient trajectories for manipulators operating in dynamic environments. An overview of the proposed trajectory planner is shown in Fig.~\ref{fig:system_overview}. The framework employs a specially designed fast trajectory planner in task space, alongside a novel model-free reinforcement learning (RL) model, to effectively address reaching tasks while navigating around obstacles. Firstly, the fast trajectory planner incorporates perception, kinodynamic path searching, and B-spline-based trajectory optimization, ensuring safe and dynamically feasible motion planning for manipulators. Secondly, we propose a novel model-free RL algorithm to solve reaching tasks while avoiding obstacles for a 6-degree-of-freedom (DoF) manipulator in complex environments. This algorithm is highly adaptable to various types of manipulators and offers superior performance compared to existing baseline RL methods. Unlike other approaches, our framework can easily adapt to new scenarios due to the use of large-scale models, eliminating the need for constructing detailed kinematic or dynamic models. Finally, comprehensive simulations and real-robot evaluations were conducted to validate the effectiveness of the proposed planner. In summary, the key contributions of this paper are as follows:

\begin{itemize}
    \item For the trajectory planner operating in the Cartesian space of the end-effector, we develop a systematic approach. This approach integrates perception, kinodynamic path searching, and B-spline-based trajectory optimization, ensuring safety and dynamic feasibility for manipulators.
    
    \item For obstacle avoidance in joint space during reaching tasks for manipulators, we propose a novel model-free RL algorithm to address reaching tasks while avoiding obstacles for a 6-DoF manipulator in complex environments, which is easily adaptable to various manipulators and offers improved performance over existing baseline RL algorithms.
    
    \item For experiments, we conduct comprehensive simulations and real-world assessments to evaluate our proposed planner. Additionally, we will make the source code of our implementation available. 
\end{itemize}

The remainder of this paper is organized as follows. Sec~\ref{sec:related_work} reviews related work on motion planning for manipulators and RL-based control methods. Sec~\ref{sec:method} presents our proposed framework, including the perception-aware trajectory planner and the modules developed to enhance PPO’s performance in trajectory tracking. Sec~\ref{sec:experiments} details the experimental setup and provides comprehensive evaluations across multiple manipulation benchmarks, along with ablation studies and comparative analyses. Sec~\ref{sec:discussion} discusses how the proposed approach addresses the previously identified challenges and limitations. Finally, Sec~\ref{sec:conclusion} concludes the paper and outlines promising directions for future research.

\section{Related Work}
\label{sec:related_work}

\subsection{Motion Planning for Manipulators}
Motion planning for manipulators, aiming to guide the end-effector along a predefined path, involves navigating safety, efficiency, and spatial constraints. This requires balancing obstacle avoidance, adherence to joint limits, and real-time execution within a limited workspace.

RRTConnect \citep{844730}, originally designed for planning movements for a human arm modeled as a $``$7-DoF kinematic chain$"$, enhances aggressiveness in the search process. E-RRT$^*$ \citep{ji2023rrt}, designed for hyper-redundant manipulators, uses ellipsoids for node linking, enhancing path planning in constrained areas. However, its specialization limits its real-time response and broader applicability, with future work aimed at improving efficiency and adaptability. PROTAMP-RRT \citep{saccuti2023protamp} introduces a unified RRT functioning in both geometric and symbolic spaces. However, its main loop, comprising task and motion planning phases, exhibits limited efficiency in manipulator obstacle avoidance. A novel bidirectional RRT$^*$ method improves obstacle avoidance in joint space for redundant manipulators, bypassing inverse kinematics \citep{dai2023novel} but only tested in static.

Probabilistic Roadmaps (PRM) sample valid configurations and link them via dynamics-compliant trajectories to build graphs for path planning. \citep{rosmann2015planning, rosmann2017integrated} initially generated a PRM or Voronoi diagram, followed by establishing a homology equivalence relation via complex analysis, applicable only in 2D. For 3D homology classes, \citep{bhattacharya2012topological} utilized electromagnetism theory, proposing a 3D homology equivalence but requiring impractical $``$genus 1$"$ obstacle decomposition as noted in \citep{jaillet2008path}, due to the difficulty of deforming 3D homotopic paths into each other. \citep{jaillet2008path} adopted a visibility deformation roadmap for a broader range of paths, but this approach, aimed at offline planning, proves too time-consuming for online applications.

Recent advancements in deep reinforcement learning (DRL) have propelled motion planning for manipulators. \citep{li2021general} developed a DRL-integrated motion planning framework for optimal path and energy solutions in kinematics, validated in both simulated and real world, but limited to static environments. \citep{zhang2021sim2real} enhanced sim-to-real adaptation and scene generalization for manipulator obstacle avoidance, Concentrating exclusively on the task space while ignoring potential obstacle navigation around the manipulator links. \citep{aljalbout2021learning} tackles vision-based obstacle avoidance for manipulators, demonstrating efficient learning of stable strategies with high success rates. However, their approach lacks reactivity in dynamic environments due to perception limitations. Unlike its extensive use in aerial drones and vehicles, training for manipulators' obstacle avoidance is confined to simulations, limited to 2D or discrete actions. Therefore, it is advantageous to integrate DRL with conventional path planners for waypoint-based path optimization. To address the aforementioned challenges, we propose creating waypoints akin to aerial drone trajectory generators and tracking trajectories utilizing RL methods, facilitating generalization and enabling obstacle avoidance for manipulators. Furthermore, the current state-of-the-art (SOTA) vision algorithm, FSA, is deployed to enhance the adaptability of the trajectory planner in complex environments.

\subsection{RL-Based Control for Manipulators}
Several research efforts have used reinforcement learning to control a robot manipulator at the joint level, \citep{ze2024h} introduced Temporal Difference Models, goal-conditioned value functions trainable via model-free learning for model-based control. Still, this method is limited to reaching tasks. To navigate towards goals while avoiding obstacles, \citep{kumar2021joint} introduced an adaptable joint-level controller through PPO. However, training with PPO is time-consuming in such tasks. Moreover, \citep{wang2021multi} enhanced PPO for training tasks, including reaching and obstacle avoidance, making it adaptable to trajectory-tracking tasks. While sufficient for space robots, the workspace for industrial robots falls short, indicating a need for algorithmic refinement. For obstacle avoidance in larger workspaces, we introduce an improved PPO that allows a 6-DoF manipulator to reach targets without collisions. This method supports low-level control within the trajectory planning framework.

Compared to reviewed manipulator motion planners, our framework divides the problem into two subproblems: trajectory planning in task space and obstacle avoidance for the manipulator's links using an RL-based joint space controller. Furthermore, we address obstacle avoidance via our proposed RL-based joint space controller, which maps task space to joint space. The waypoints generated by the trajectory planner ensure a safe and efficient trajectory in task space.

\section{Method}
\label{sec:method}
We aim to enhance the capability of manipulators for safe and efficient motion planning in environments with both static and dynamic obstacles. The primary contribution is the development of an integrated vision-based trajectory planner coupled with an enhanced RL joint space controller, enabling manipulators to achieve goal-oriented motion with obstacle avoidance. In this section, we briefly discuss vision-based trajectory planning in Sec~\ref{sec:method_1}. We then discuss in detail the construction of an improved PPO and its role in enhancing the performance of the reaching task with obstacle avoidance for manipulators in Sec~\ref{sec:method_2}. Finally, in Sec~\ref{sec:method_3}, we discuss combining these elements to develop a fast trajectory planner.

\subsection{Perception-Aware Trajectory Planning for Manipulators}
\label{sec:method_1}

\begin{figure*}[!t]
      \centering
      \includegraphics[width=1.0\linewidth]{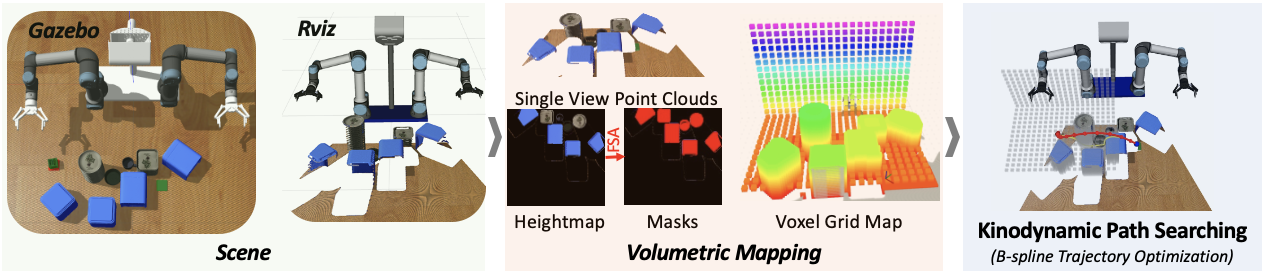}
      \caption{\textbf{Illustration of perception-aware Trajectory Planning}: Using the FSA, object masks are merged with point clouds to create a Voxel Grid Map, providing a detailed basis for improved navigation and planning decisions.}
      \label{fig:illustration}
\end{figure*}

Single-view point clouds often suffer from incompleteness. To attain 3D point cloud completion, while the use of multiple sensors yields a detailed composite, it significantly increases both cost and system complexity. To streamline this process, we leverage SOTA object detection to facilitate the generation of complete point clouds. The Segment Anything Model (SAM) revolutionizes object segmentation but faces deployment challenges in real-time applications due to its high computational needs. \citep{zhao2023fast} introduced an alternative to the Segment Anything Model, 50 times faster and suitable for robotic perception, leveraging FSA to aid in the creation of voxel grid maps for environments.  In detail, consider a daily life scene with various bottles and boxes on a table. A Kinect sensor captures images, converting them into a height map. FSA could detect masks for objects from the height map of the scene. To eliminate the noisy background, we filter all masks based on their $z$ position to isolate objects above the table, effectively removing the background and shadows. These filtered masks and table information help reconstruct the occluded sections of the point cloud of the scene. An overview of this process is shown in Fig.~\ref{fig:illustration}.

\begin{figure}[!t]
      \centering
      \includegraphics[width=1.0\linewidth]{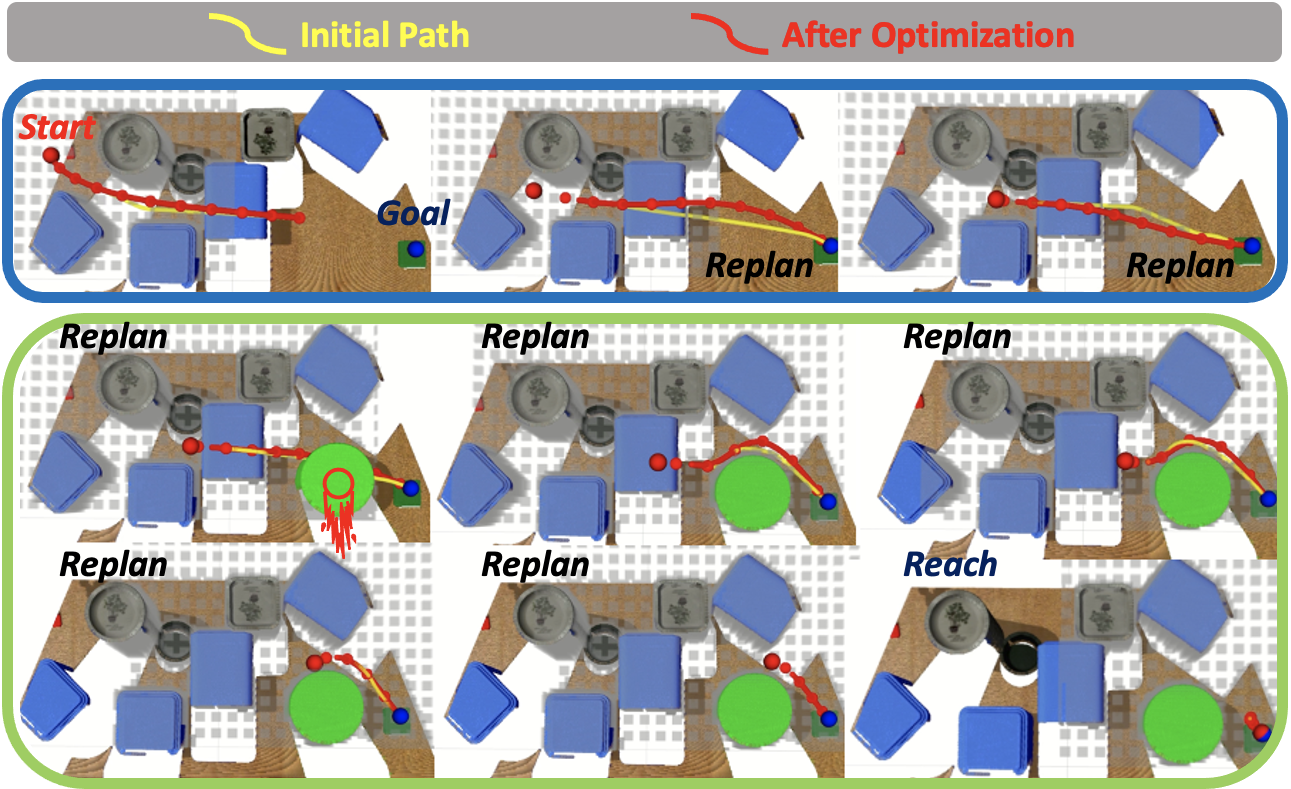}
      \caption{\textbf{Trajectory Planning Process}: This example illustrates trajectory planning in a dynamic environment, where the yellow trajectory signifies initial planning. The start and target points are marked by a red and a blue ball, respectively. Upon encountering an unforeseen obstacle, represented by a green cylinder, the system triggers a replanning mechanism. Initially, the path does not circumvent the obstacle due to outdated map data. However, once the obstacle is detected, the system generates a new trajectory to navigate around it. This revised path, highlighted in red with waypoints, serves as input for the RL-based controller, ensuring adaptive and efficient navigation}
      \label{fig:trajectory}
\end{figure}

After estimating the full point cloud of the scene, we utilize them for real-time trajectory planning in complex 3D environments. Drawing inspiration from SOTA drone path planning methods \citep{zhou2019robust, zhou2020robust, zhou2021raptor}, we aim to meet the requirements of real-time performance. Our approach includes kinodynamic path planning combined with B-spline-based trajectory optimization to create a trajectory that is both safe and efficient. Additionally, the manipulator frequently needs to replan its trajectory due to unknown environments. This replanning occurs under two conditions: firstly, when the current path intersects with newly detected obstacles, ensuring immediate access to a safe trajectory; and secondly, at regular time intervals, which allows for continuous updates to the trajectory based on the latest environmental map. Fig.~\ref{fig:trajectory} illustrates an example of navigating a manipulator from a starting point to a goal while avoiding both static and dynamic obstacles.

\subsection{RL-based Joint Space Controller for Obstacle Avoidance} 
\label{sec:method_2}
Our goal is to develop a neural network-based controller for precise trajectory tracking, which, through extensive training, can map task and joint spaces without calculating inverse kinematics or dynamics. This learned controller demonstrates a significant advantage in navigating obstacles and executing maneuvers by directly mapping the current state $s_t$ to an action $a_t$ through a policy $\pi_\theta$. This policy, optimized to maximize rewards, signifies a shift from traditional control strategies. In our framework, forward kinematics (FK) acts as the interface between joint-space control and Cartesian task objectives, supporting state representation by computing the current end-effector pose for the observation vector, reward shaping by evaluating pose error relative to the target, and task evaluation by verifying at test time whether the final pose satisfies the required tolerance. (detailed in the appendix~\ref{appendix})

\subsubsection{State and Action Representations}
Many researchers often use task space representations for state and action in robotic manipulator tasks, which can inadequately address link-obstacle collisions. In our method, state representation includes positions of the 6 joints, the pose of the end-effector, the pose of the target point, the pose error between the target and end-effector, and distances between obstacles and manipulator links (excluding the base link), resulting in a 25-dimensional state vector \( \bm{s} \in \mathbb{R}^{25} \). For action selection, desired joint velocities \(\dot{\bm{q}}_d\) are determined and subsequently integrated to calculate the desired joint angles \(\bm{q}_d\). These angles serve as inputs to a proportional-derivative (PD) controller, which then generates the necessary torques for controlling the robot. The state is represented as:
\begin{equation}
    \bm s_t = <\bm q_t, \bm p_e, \bm p_t, \bm {error}, \bm d_{obs}>
    \label{equ:1}
\end{equation} 
where \(\bm{q}_t = (q_{t1}, \ldots, q_{t6})\) denotes the positions of the 6 joints, \(\bm{p}_e = (p_{ex}, p_{ey}, p_{ez}, o_{e})\) and \(\bm{p}_t = (p_{tx}, p_{ty}, p_{tz}, o_{t})\) represent the poses of the end-effector and target, respectively. These poses can be defined using either Euler angles or quaternions, resulting in a length of either 6 or 7; in this paper, quaternions are utilized. The \(\bm{error}\) vector quantifies the discrepancy between the end-effector and target, incorporating the sum of distance errors from three points delineated along orthogonal axes and orientation error, as expounded in \citep{molchanov2019sim, zhou2019continuity}. \(\bm{d}_{obs}\) signifies the minimal distance between obstacles and the manipulator's links. This distance calculation is reformulated as a geometric problem to ascertain the nearest point in space to various links (detailed in the appendix~\ref{appendix}). The action space is represented by a vector, $\bm a_t = <\dot{\bm{q}}_d>$, where $\dot{\bm{q}}_d = (\dot{q}_{d1} \dots \dot{q}_{d6})$ denotes the velocity of the 6 joints.

\subsubsection{Strategy for Learning}
In this paper, we adopt PPO \citep{schulman2017proximal}, an advanced reinforcement learning method praised for its balance between stability and efficiency, especially in complex robotic tasks with obstacle avoidance. PPO uses a policy gradient approach, enhancing training stability through a clipped surrogate objective function that limits policy changes, thus optimizing rewards efficiently:
\begin{equation}
   \begin{aligned}
    \bm{L}^{\bm{CLIP}}(\theta^{\pi})&=\mathbb{\dot{E}}_t[min(r_t(\theta^{\pi})\hat{A_t},\\
    &\quad clip(r_t(\theta^{\pi}),1-\epsilon,1+\epsilon)\hat{A_t})]
   \end{aligned}
\end{equation}
\begin{equation}
    \hat{A_t} = \delta_t+(\gamma\lambda)\delta_{t+1}+\cdots+\cdots+(\gamma\lambda)^{T-t+1}\delta_{T-1}
\end{equation}
\begin{equation}
    \delta_t = r_t+\gamma V(s_{t+1})-V(s_t)
\end{equation}
\begin{equation}
    V(s) = \mathbb{E}_{s,a\sim\pi}[G(s)|s]
\end{equation}
\begin{equation}
    G(s) = \sum_{i=t}^{\infty}\gamma ^ {i-t}r(s_i)
\end{equation}
where \(\theta^{\pi}\) indicates the policy network's parameters, with \(r_t(\theta^{\pi})\) as the probability ratio. The Generalized Advantage Estimator (GAE), \(\hat{A_t}\), calculates the policy gradient. Rewards at time \(t\) are noted as \(r_t\), with a clipping parameter \(\epsilon=0.2\) and discount factor \(\gamma=0.99\). \(V(s)\) represents the expected return from state \(s\), and \(G\) the discounted rewards. \(V_{target}(s)\) serves as the target value for the value loss function, streamlining optimization. The value loss function is expressed as follows:
\begin{equation}
    \bm{L}_V(\theta^{V})=\mathbb{E}_{s,a\sim\pi}[(V(s)-V_{target}(s))^ 2]
\end{equation}

Facing challenges like time-consuming training and suboptimal results with PPO in complex robotic tasks, we introduce two enhancements to improve its performance.

\paragraph{Action Ensembles}
The established principle that ensembles enhance stability, reduce variance, and improve uncertainty estimation is well recognized. In RL, they are especially valuable in offline settings, where robust uncertainty modeling is essential for safe, effective policy learning. \citep{ghasemipour2022so, an2021uncertainty, lee2022offline} As a method grounded in policy gradients, PPO strives to approximate the optimal policy through its objective function. Furthermore, Gaussian distributions are favored for modeling optimal policies within the domain of robotics tasks, as they accurately capture the intricacies of real-world dynamics. For the manipulator reaching task, actions are modeled as $\pi_\theta(a_t|s_t) \sim \mathcal{N}(\mu_\theta(s_t), \sigma^2_\theta)$, where $\sigma^2_\theta$ represents the uncertainty and variability of actions. By utilizing KL divergence in the PPO framework to quantify the disparity between the optimal and current policies, the ensuing objective function is derived \citep{schulman2017proximal}:
\begin{equation}
    \min_{\theta} D_{\text{KL}} \left( \pi^{*}(\cdot|s_t) || \pi_{\theta}(a_t|s_t) \right)
\end{equation}

\begin{figure}[!t]
    \includegraphics[width=1.0\linewidth]{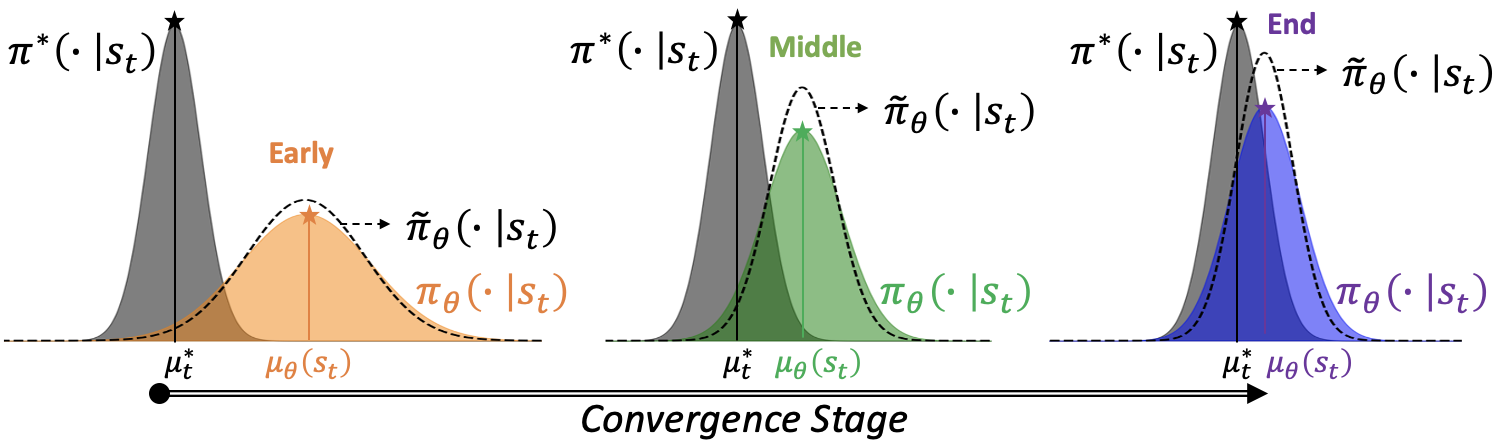}
\caption{The policy approximation towards the optimal policy in a 1-dimensional example unfolds in three stages from left to right: early exploration, middle exploration, and later convergence. $\pi^{*}(\cdot|s_t)$, $\pi_\theta(\cdot|s_t)$, and $\widetilde{\pi}_\theta(\cdot|s_t)$ represent the optimal, current, and adjustable policies, respectively.}
\label{fig:policy_approximation}
\end{figure}

To optimize the objective function, the ideal $\theta^*$ should adhere to the conditions $\mu^*(s_t) = \mu_\theta(s_t)$ and $\Sigma^* = \Sigma_\theta$. Typically, the optimal action $a^*_t$ at any given time $t$ is deterministic with minimal uncertainty. Therefore, the reduction of policy uncertainty, denoted as $\Sigma_\theta$, alongside the attainment of the optimal $\mu^*(s_t)$, holds paramount importance. The mitigation of uncertainty frequently entails the aggregation of multiple action values, predicated on the principle that averaging samples drawn from a Gaussian distribution effectively diminishes policy uncertainty by smoothing the distribution of action outputs. Nevertheless, this approach could potentially inhibit early exploration and pose the risk of local optima stagnation. 
To effectively balance exploration and convergence, we vary sample sizes across distributions\citep{wang2021multi}, demonstrating the process in Fig.~\ref{fig:policy_approximation} with a one-dimensional toy example. These enhancements are integrated into our refined action ensembles method as specified in the formula below:
\begin{equation}
    \bm a_{t,j}\sim N(\mu_\theta(s_t), \delta_\theta),\quad \bm a_t = \underset{j}{mean}(\bm a_{t,j})
\end{equation}
where $\bm a_{t,j}$ represents $j^{th}$ action at time t, with $j$ varying from 1 up to but not including $i$. $e_n$ and $e_a$ represent the counts of the current and total episodes, respectively. The number of samples, denoted as $i$, varies based on the chosen distribution. We adjust sample sizes for different distributions and detail their implementation within the PPO framework below.

\begin{itemize}
  \item AE with Linear (AEL): Linear improvement for sample selection, defined as follows:
\begin{equation}
    i = 1+ \alpha \dfrac{e_n}{e_a}
\end{equation}
$\alpha$ acts as a hyperparameter that significantly influences the convergence rate.

  \item AE with Poisson Distribution (AEP): Poisson distribution for sample selection, defined as follows:
\begin{equation}
    i\sim clip(Poisson(\beta), 1, \beta),\quad \beta = 1+ \alpha \dfrac{e_n}{e_a}
\end{equation}
$\beta$ denotes the parameter of the Poisson distribution.

  \item AE with Beta Distribution (AEB): Beta distribution for sample selection, defined as follows:
\begin{equation}
    i\sim clip(Beta(a, b), 1, a) 
\end{equation}
\begin{equation}
    a = 1+ \alpha \dfrac{e_n}{e_a} \quad b = 1+ \beta \dfrac{e_n}{e_a} 
\end{equation}
where $a$ and $b$ represent the parameters of the Beta distribution. $\alpha$ and $\beta$ are the hyperparameters influencing the convergence rate.
  
  \item AE with Exponential Distribution (AEE): Exponential distribution for sample selection, defined as follows:
\begin{equation}
    i\sim clip(Exp(\lambda), 1, \lambda),\quad \lambda = 1/(1 + \alpha \dfrac{e_n}{e_a})
\end{equation}
$\lambda$ is the parameter of the Exponential distribution.

  \item AE with Weibull Distribution (AEW): Weibull distribution for sample selection, defined as follows:
\begin{equation}
    i\sim clip(Weibull(k, \lambda), 1, \lambda)
\end{equation}
\begin{equation}
    k = 1+ \alpha \dfrac{e_n}{e_a} \quad \lambda = 1+ \beta \dfrac{e_n}{e_a}
\end{equation}
where $k$ and $\lambda$ represent the parameters of the Weibull distribution. $\alpha$ and $\beta$ are the hyper-parameters influencing the convergence rate.
\end{itemize}

Following experimentation, AEW is selected as the augmentation method for enhancing PPO in our task. Details regarding the parameter selection are included in the appendix~\ref{appendix}.

\paragraph{Policy Feedback}

\begin{figure}[!b]
      \centering
      \includegraphics[width=1.0\linewidth]{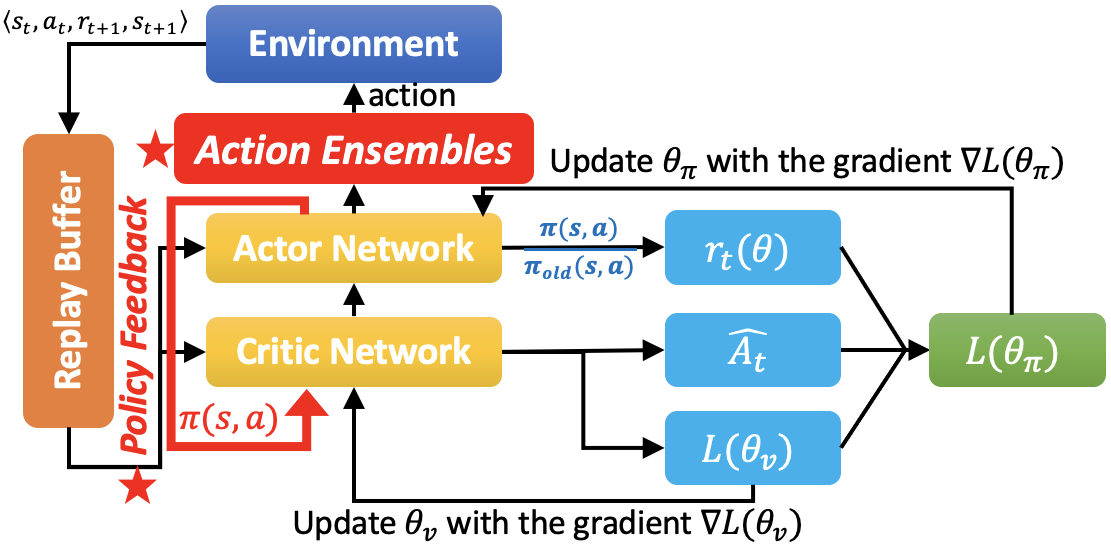}
      \caption{\textbf{The Framework of the PPO with Enhancements}: The action ensembles (\textit{shown in red block}) and the policy feedback (\textit{red line})}
      \label{fig:framework_ppo}
\end{figure}

Within the PPO framework featuring an Actor-Critic (AC) architecture, the critic network refines the actor's policy by estimating the value function. Nonetheless, the policy's indirect influence on value function updates may heighten instability in DRL algorithms, highlighting the challenge of maintaining equilibrium between actor and critic adaptations for stable learning. The stability of PPO is enhanced by directly incorporating the policy into the updates of the value function\citep{gu2021proximal}. This approach enables the critic to rapidly adapt to shifts in policy. Furthermore, \citep{gan2024reflective} introduced Reflective Policy Optimization (RPO), which incorporates both past and future policy states into the critic to promote more stable learning dynamics. Such enhancement is facilitated through the implementation of an adaptive clipped discount factor, which allows for a more dynamic adjustment in response to policy changes, thereby promoting a more robust and stable learning environment within the framework.
\begin{equation}
    \gamma(s,a;\eta) = clip(\pi(s,a), \eta, 1) \quad \eta\in(0.6, 0.99)
    \label{equ:17}
\end{equation}
in which $\pi(s,a)$ represents the policy and the adaptive $\gamma$ can join in the update of critic network. Both the AC neural network consists of 3 layers, where each layer consists of $256$ neurons. The first 2 layers use the $tanh$ activation function. The only distinction between actor and critic networks is that the critic network generates only a single scalar value, while the actor produces a vector of 6 values. For both networks, we utilize the Adam optimizer. Fig.~\ref{fig:framework_ppo} shows the overall architecture of the improved PPO.

\subsubsection{Reward Function}
The following functions outline the approach utilized for calculating the reward:
\begin{equation}
    \bm{R}(s,a)= -[\omega_1e^2+\ln{(e^2+\tau)}+\omega_2\sum_{i=1}^{n}\psi_i]
    \label{equ:2}
\end{equation}
\begin{equation}
    \psi_i=max(0, 1-||d_i||/d_{max})
    \label{equ:3}
\end{equation}
where \(e= \sum \bm{error}\) quantifies the pose discrepancy between the target and the end-effector. The use of the \(\ln(\cdot)\) aims to minimize error \(e\) while maximizing the reward; \(\psi_i\) denotes penalties for collisions. The coefficients \(\omega_1\), \(\omega_2\), and \(\tau\), along with \(d_{max}\) which defines the error threshold between the end-effector and target pose, are determined through trial and error to be \(\omega_1 = 10^{-3}\), \(\omega_2 = 0.1\), \(d_{max} = 0.08\), and \(\tau = 10^{-4}\).

\subsubsection{Improved PPO}
According to previous advancements, the loss functions of AC networks can be represented as follows:

\begin{equation}
   \begin{aligned}
    \bm{L}(\theta^{\pi})&=\mathbb{\dot{E}}_t[min(r_t(\theta^{\pi}){A_t}, \\
   &\quad clip(r_t(\theta^{\pi}),1-\epsilon,1+\epsilon){A_t})]
   \end{aligned}
   \label{equ20}
\end{equation}

\begin{equation}
    \bm{L}(\theta^{V})=\mathbb{\dot{E}}_t[(R_t^\pi-V_t)^2]
    \label{equ21}
\end{equation}

\begin{figure}[!b]
  \centering
  \includegraphics[width=1.0\linewidth]{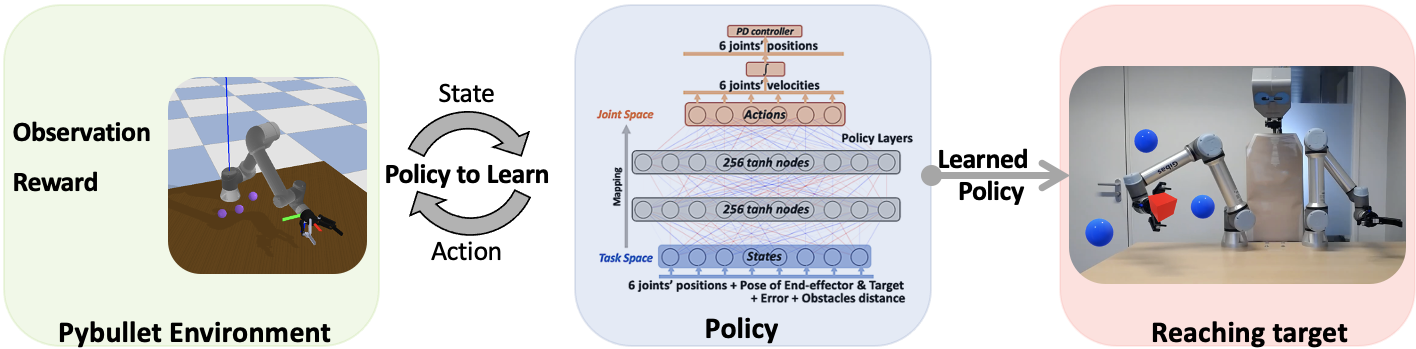}
  \caption{\textbf{Learning Strategy Deployment Overview:} The agent is first trained in Pybullet to learn the task-to-joint space mapping of the UR5e manipulator. This trained policy is then adapted from one simulation to another (Sim-to-Sim) in Gazebo and applied directly to real-world robot scenarios without further fine-tuning.}
  \label{fig:deployment}
\end{figure}

Fig.~\ref{fig:framework_ppo} shows the overall architecture of the improved PPO, and Algorithm ~\ref{alg1} represents the pseudocode of the improved PPO. By combining all the proposed strategies together, the entire process for the robot to avoid obstacles while reaching the goal is summarized in Algorithm ~\ref{alg2} and Fig.~\ref{fig:deployment}.

\begin{algorithm}[!t]
    \caption{Improved PPO Pseudocode}
    \label{alg1}
    \begin{algorithmic}
      \STATE Orthogonally initialize the actor and critic networks
      \STATE Initialize optimizer as Adam with learning rates
      \STATE Set $\lambda$, $a_{max} = 3.14$, clip parameter: $\epsilon$
      \STATE Set AEW parameter: $\alpha=7$, $\beta=20$
      \STATE Set Policy Feedback parameter: $\eta \in [0.6, 0.99]$
      
      \FOR{iteration = 1 to max\_steps}
      
        \STATE Get $s_t$ from the environment
        \STATE Sample $a_t$ from the baseline policy $\pi_{b\theta}(a_t|s_t)$
        \STATE Use $\bm{a_t} \sim AEW(\mu_\theta(s_t), \delta_\theta)$ to publish $\bm{a_t}$
        \STATE Execute $a_t$, get $<s_t, a_t, \pi_\theta, r_t, s_{t+1}>$ and store them into buffer $D$
        \STATE Calculate clipped discount factor $\gamma(s,a;\eta)$ by Eq.~\ref{equ:17}
        \STATE Compute reward: $R_t^\pi = \sum \limits_{i=t}^{T} r_i \prod \limits_{j=t}^n \gamma(s,a;\eta)$
        \STATE Compute advantage function: $A_t^\pi = r_t + \gamma V_{t+1}^\pi - V_t^\pi$
        \STATE Use Eq.~\ref{equ20} to accumulate gradients with respect to $\theta_\pi$
        \STATE Use Eq.~\ref{equ21} to accumulate gradients with respect to $\theta_V$
      \ENDFOR
    \end{algorithmic}
\end{algorithm}

\begin{algorithm}[!t]
    \caption{Manipulator Control via Improved PPO}
    \label{alg2}
    \begin{algorithmic}
        \FOR{number of epochs}
            \STATE Set \texttt{i\_epoch} as random seed
            \STATE Orthogonally initialize the actor and critic networks
            \STATE Set $\lambda$, maximum action, learning rates, and clip parameter $\epsilon$
            \FOR{number of episodes}
                \STATE Set a target position $p_t$ randomly
                \WHILE{True}
                    \STATE Get $s_t$ from the environment
                    \STATE Choose the action using $\bm{a_t} \sim \mathcal{N}(\mu_\theta(s_t), \delta_\theta)$
                    \STATE Clip $\bm{a_t}$ to ensure safety for the environment
                    \STATE Execute $a_t$, get $<s_t, a_t, \pi_\theta, r_t, s_{t+1}>$ and store them into buffer $D$
                \ENDWHILE
            \ENDFOR
            \FOR{every time step $t$}
                \FOR{$k$ epochs}
                    \STATE Select a minibatch $b_k$ in $D$
                    \STATE Utilize Algorithm 1 to update networks
                \ENDFOR
            \ENDFOR
        \ENDFOR
    \end{algorithmic}
\end{algorithm}

\subsection{Fast Trajectory Planner}
\label{sec:method_3}

To achieve efficient motion planning for manipulators, we combine the perception-aware trajectory planning described in Sec.\ref{sec:method_1} with the RL-based joint space controller from Sec.\ref{sec:method_2}, creating a fast and adaptive trajectory planner capable of handling complex environments with both static and dynamic obstacles. The key innovation of this fast trajectory planner is its ability to replan on-the-fly. As the manipulator navigates through its environment, the system continuously monitors for new obstacles. If an obstacle is detected, the planner immediately triggers a replanning mechanism. The perception-aware module updates the voxel grid, and a new safe trajectory is generated using the B-spline optimization technique. This updated path is then fed into the RL-based controller, which ensures smooth and efficient navigation around the obstacle.

\begin{figure}[!b]
  \centering
  \includegraphics[width=1.0\linewidth]{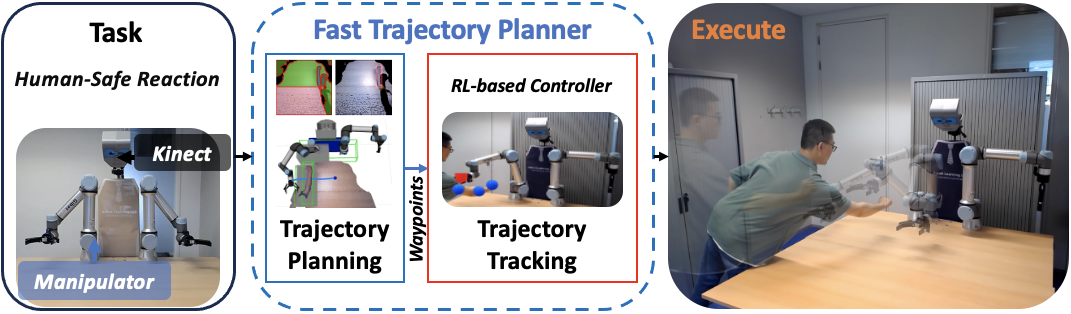}
  \caption{\textbf{Fast Trajectory Planner:} This diagram illustrates the workflow of the fast trajectory planner, which combines real-time perception, B-spline-based trajectory optimization, and RL-based control. The system dynamically replans trajectories in response to changes, ensuring safe and efficient navigation during reactive tasks.}
  \label{fig:ftp}
\end{figure}

Moreover, to enhance overall system efficiency, the planner integrates a kinodynamic path-searching algorithm, which takes into account the manipulator’s dynamic and kinematic constraints. This ensures that the trajectories not only avoid obstacles but also respect the physical limits of the manipulator. In summary, the fast trajectory planner seamlessly integrates real-time perception, dynamic path planning, and reinforcement learning-based control to enable safe, adaptive, and efficient motion planning for manipulators in complex environments. This allows manipulators to achieve goal-oriented tasks while autonomously avoiding both static and dynamic obstacles, as demonstrated in Fig.~\ref{fig:ftp}.

\section{Experiments}
\label{sec:experiments}
In this section, we evaluated our methods in both simulations and the real world. Our experiments demonstrated that: 1) The improved PPO surpassed other baselines in performance and robustness across different simulation platforms. 2) The RL-based controller achieved precise and stable joint-level control of a robot manipulator. 3) The fast trajectory planner embedded within the proposed controller facilitated real-time trajectory planning for the manipulator.

\subsection{Experimental Setups}
We conducted experiments on the following tasks:

\begin{itemize}
    \item \textbf{Reaching with Obstacle Avoidance:} The enhanced PPO was trained in Pybullet to expedite training, with the model tested in Gazebo to validate the robustness.
    \item \textbf{Motion Planning in Static Obstructed Environments:} The manipulator navigated from start to goal, avoiding obstacles in varied environments.
    \item \textbf{Motion Planning in Moving Obstacles Environments:} The manipulator avoided moving obstacles from start to goal in dynamic environments.
    \item \textbf{Reaching the Moving Target with Obstacle Avoidance:} The manipulator achieved obstacle avoidance while reaching a moving target.
    \item \textbf{Generalization Across Various Manipulators:} The enhanced PPO algorithm was evaluated on manipulators with different DoF to validate its generalization capability for the reaching task.
\end{itemize}

Our experiments utilized a 6-DoF UR5e manipulator, conducted on a PC equipped with Ubuntu 20.04, an Intel Xeon i7 processor at 3.20 GHz, and an NVIDIA RTX 2080 Ti card.

\subsection{Strategy Evaluation in Reaching Tasks}
The experimental workspace was shaped as a sphere with an 85 cm radius, excluding a 30 cm radius cylinder at the base of the manipulator. During training, targets were randomly placed within the workspace, with three 5 cm-diameter spherical obstacles located near the target or the manipulator's links, serving as areas to avoid (detailed in the appendix~\ref{appendix}). To assess the efficiency of the proposed RL method, we investigated the impact of each algorithm component, compared strategy performance with baselines, and tested strategy robustness. All experiments involved running \(1.5 \times 10^4\) episodes, each limited to 100-time steps, totaling 1.5 million steps. To minimize serendipity, we consistently ran the algorithms five times with unique random seeds for all experiments.

\subsubsection{Ablation Experiments}

In this round of experiments, we systematically modify various components of our model to assess their impacts on performance.

\begin{figure*}[!t]
    \includegraphics[width=0.32\linewidth]{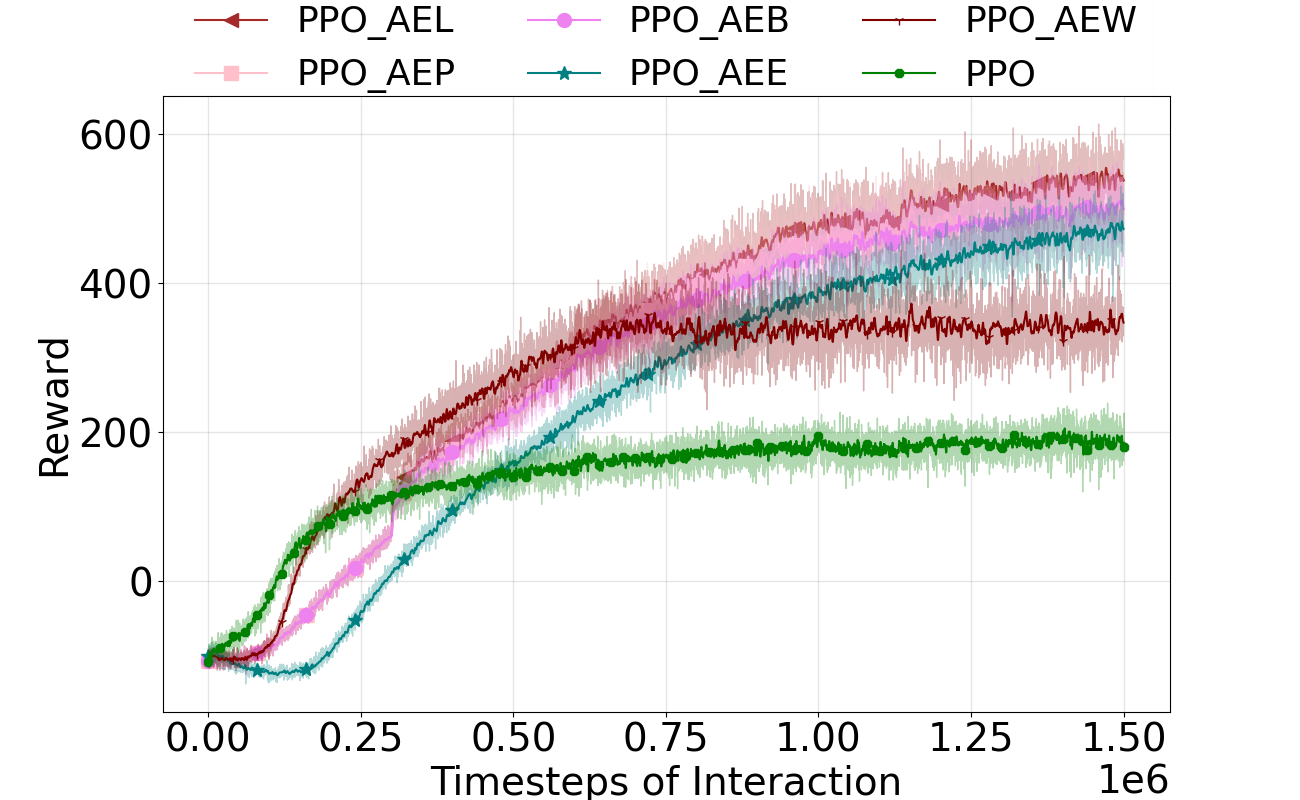}
    \includegraphics[width=0.32\linewidth]{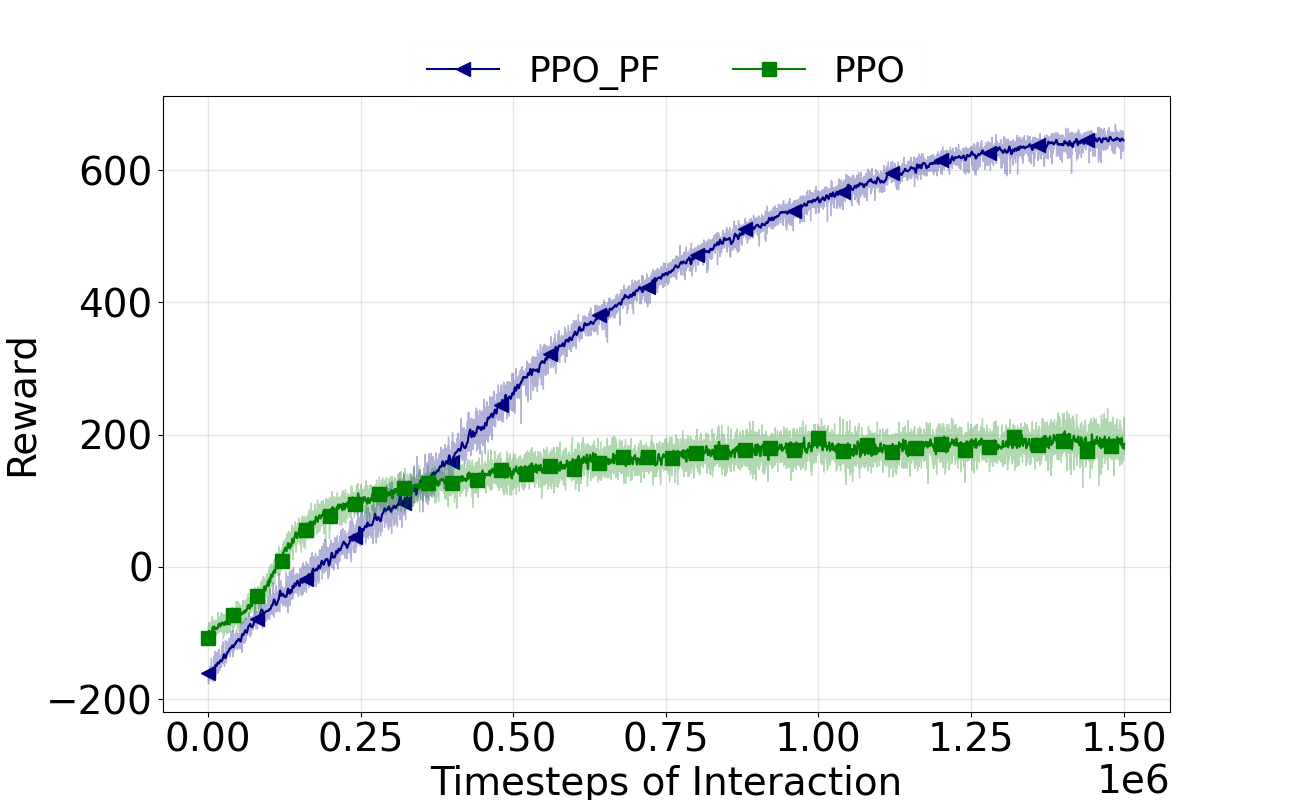}
    \includegraphics[width=0.32\linewidth]{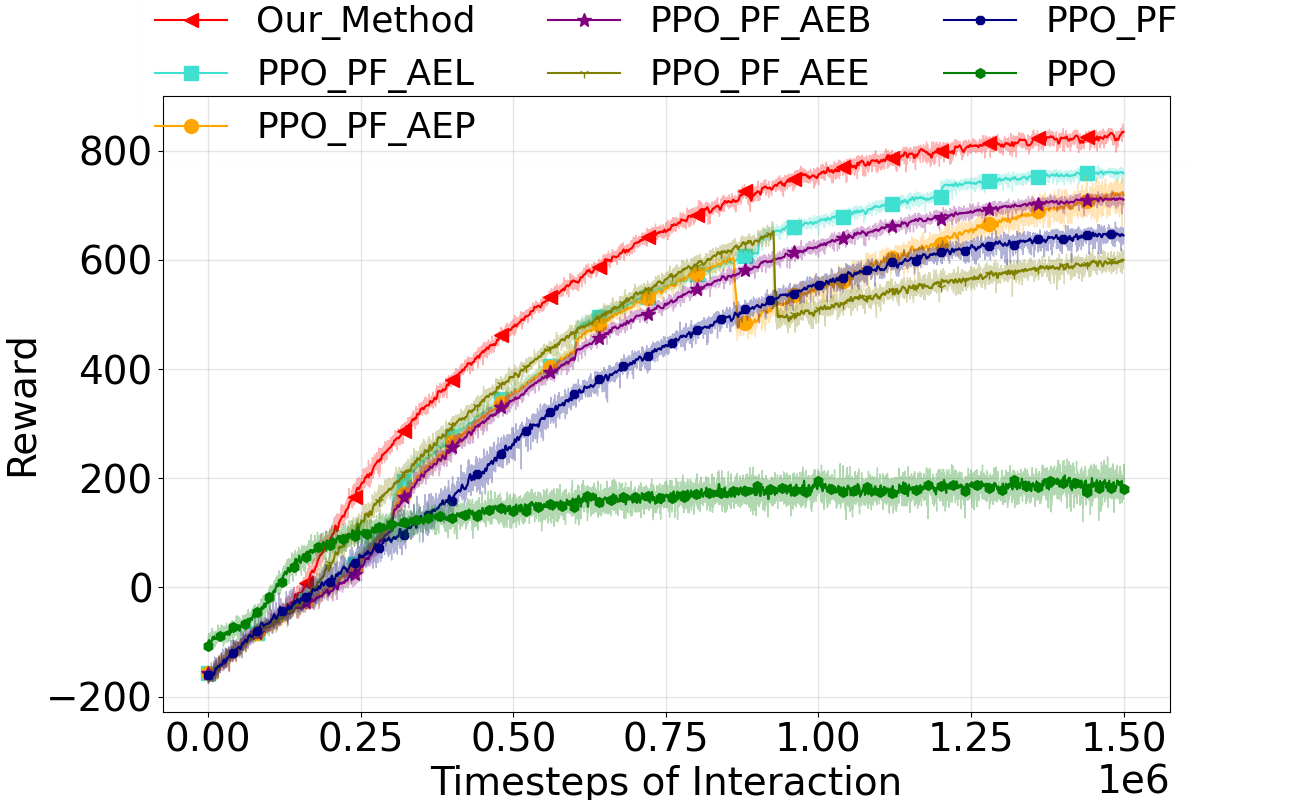}
    \par\noindent\makebox[0.32\linewidth][c]{\footnotesize(a)} 
    \makebox[0.32\linewidth][c]{\footnotesize(b)} 
    \makebox[0.32\linewidth][c]{\footnotesize(c)} 
    \vspace{-3mm}
\caption{\textbf{Summary of Ablation Experiments}: (a) Accumulated reward comparison for PPO with various AE methods across 5 random seeds; (b) PPO with PF method compared using 5 random seeds; (c) PPO performance with dual enhancements over 5 seeds. Variance is depicted by shaded areas.}
\vspace{-3mm}
\label{fig:ablation}
\end{figure*}

\paragraph{Action Ensembles}
Our evaluation of AE's impact on convergence is presented in Fig.~\ref{fig:ablation} (a), which illustrates the convergence rates and optimal rewards for various AE methods across different distributions. Fig.~\ref{fig:ablation} (a) demonstrates that PPO, when augmented with AE, achieves higher rewards, indicating enhanced accuracy in the reaching task with obstacle avoidance. The AEL variant secures the highest reward across all tested distributions. Additionally, the Weibull distribution exhibits rapid improvement in the initial phase of the convergence process, even though its total accumulated rewards may not be the highest. Furthermore, this suggests that AE enhancements are effective for robotic tasks. To identify the optimal approach for various applications, conducting experiments tailored to each specific task is essential. Varying parameters across different distributions can influence outcomes. To explore this, we conducted extensive ablation studies (detailed in the appendix~\ref{appendix}) that examined how changes in distribution parameters affect the mean value intervals. These studies indicate that the AEW method's performance is largely stable despite these variations.

\paragraph{Policy Feedback}
Fig.~\ref{fig:ablation} (b) demonstrates that incorporating PF into PPO results in higher rewards compared to the original PPO. From Fig.~\ref{fig:ablation} (c), it is evident that combining AE and PF enhancements significantly improves reward outcomes for tasks involving obstacle avoidance. Specifically, the PPO\_PF\_AEW variant achieves the highest accumulated rewards, as described in Sec~\ref{sec:method_2}. The AE component addresses the challenges of inhibited early exploration and the risk of local optima stagnation found in the original PPO, effectively balancing exploration and convergence. The PF component, implemented through an adaptive clipped discount factor, allows for dynamic adjustment in response to policy changes, fostering a more robust and stable learning environment within the framework. These results validate our theoretical approach.

\subsubsection{Comparison Experiments}
\begin{figure*}[!t]
    \includegraphics[width=0.32\linewidth]{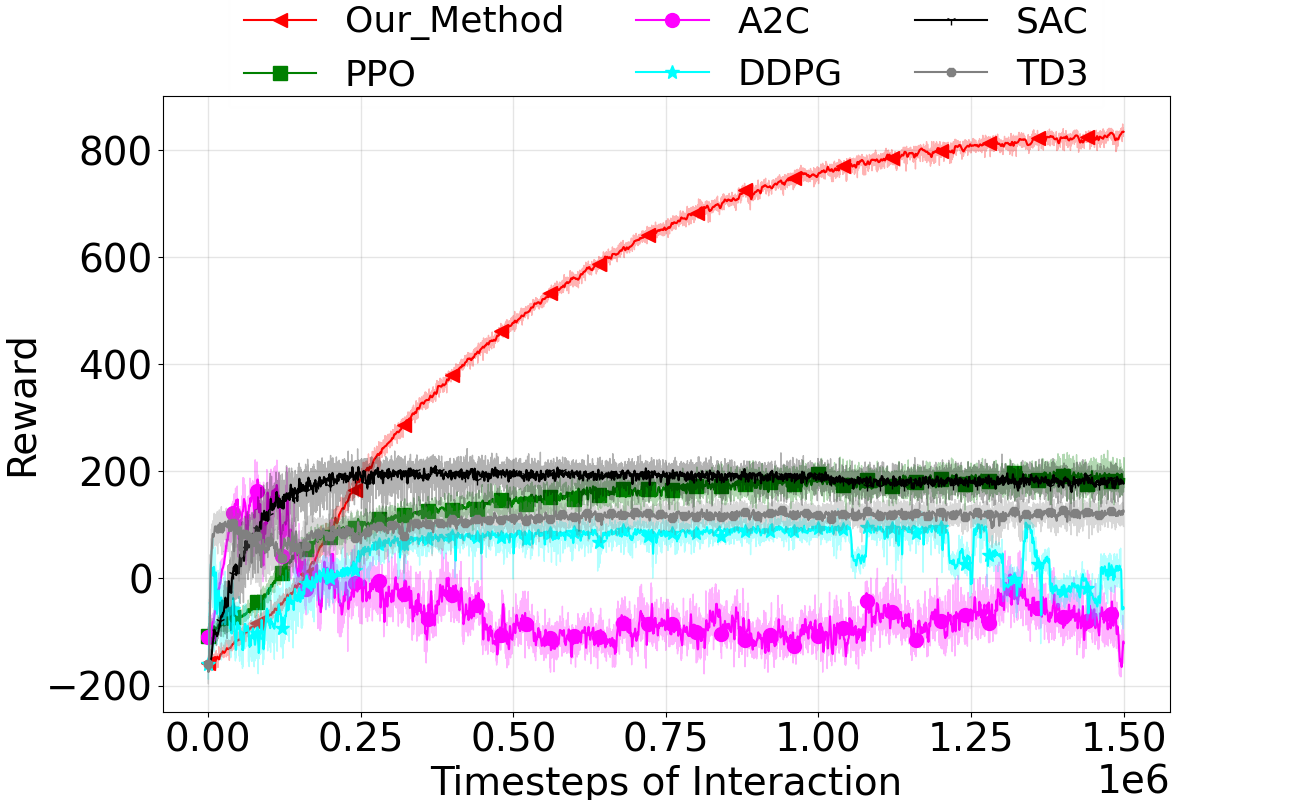}
    \includegraphics[width=0.32\linewidth]{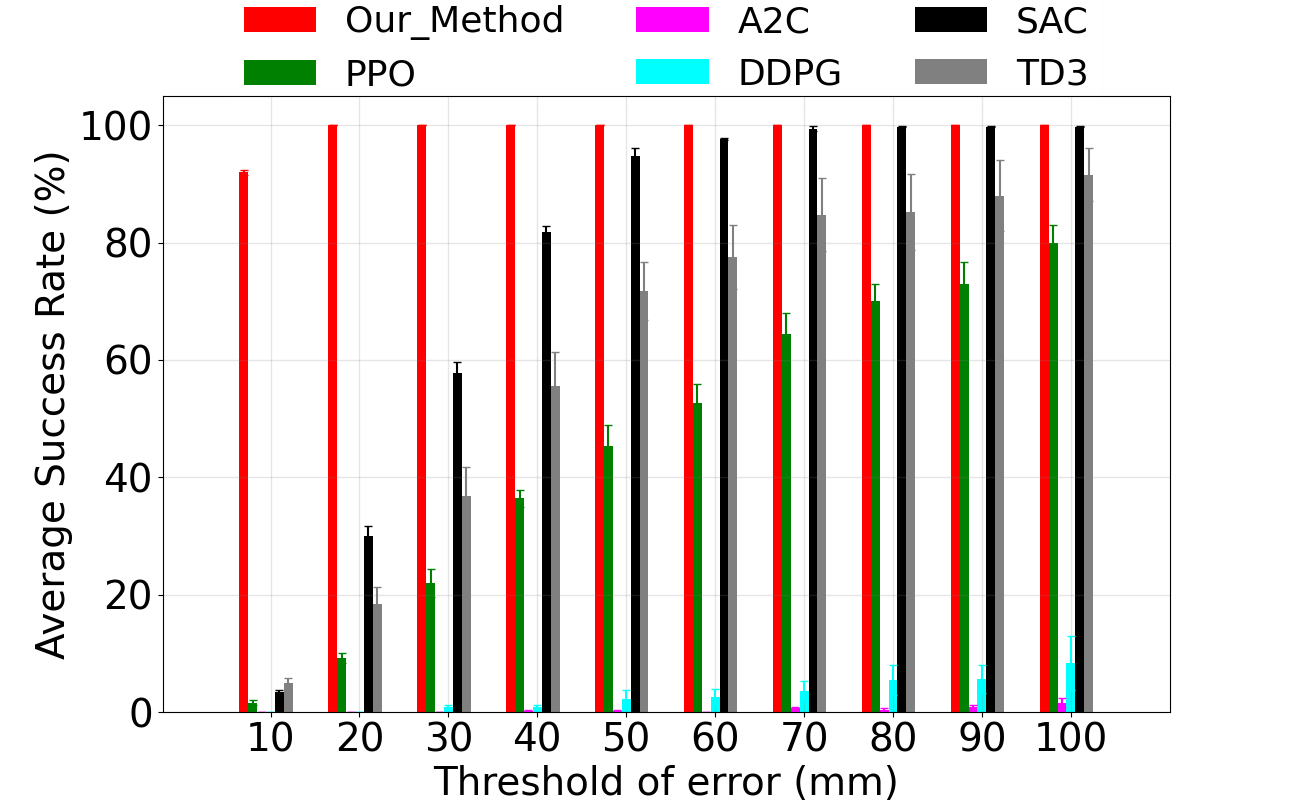}
    \includegraphics[width=0.32\linewidth]{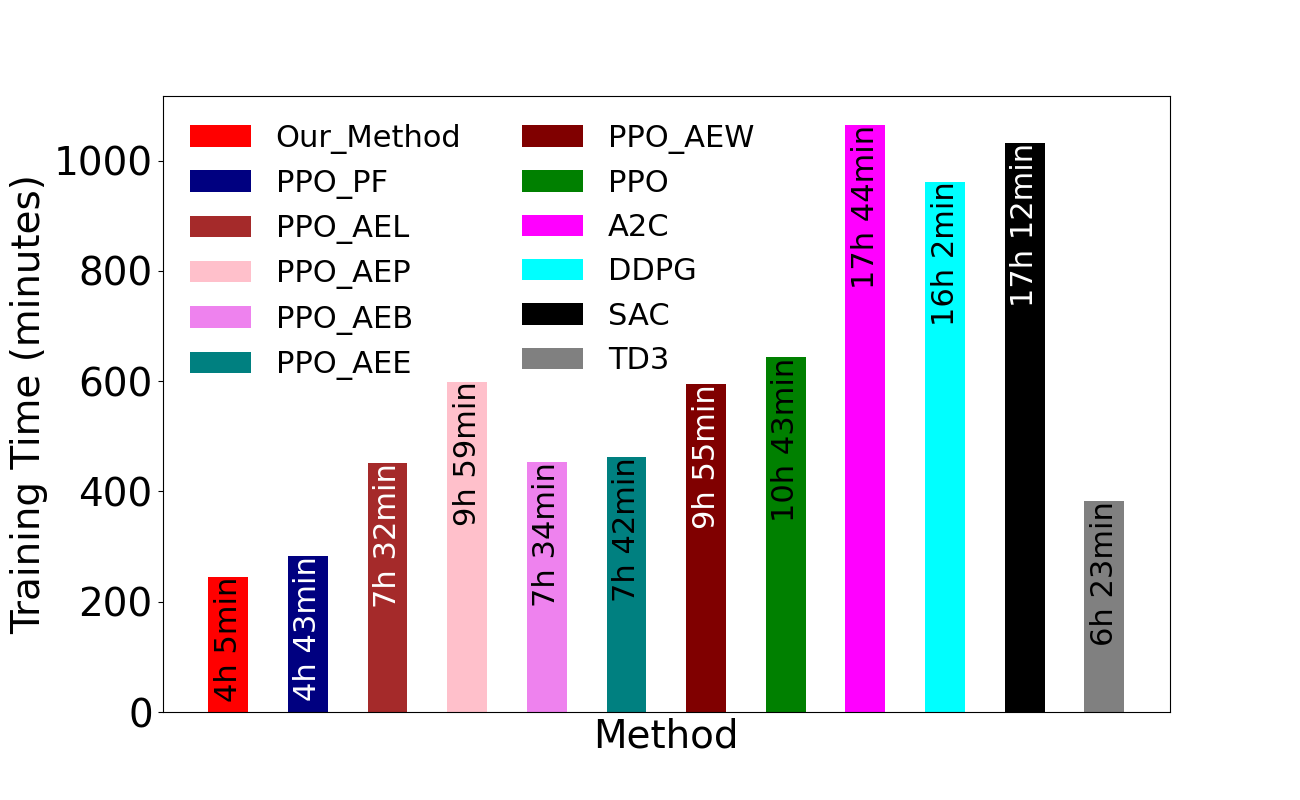}
    \par\noindent\makebox[0.32\linewidth][c]{\footnotesize(a)} 
    \makebox[0.32\linewidth][c]{\footnotesize(b)} 
    \makebox[0.32\linewidth][c]{\footnotesize(c)} 
    \vspace{-3mm}
\caption{\textbf{Summary of Comparison Experiments}: (a) Comparison result of accumulated reward utilizing 5 random seeds for our method and other baselines; (b) Comparison of success rate on reaching task with obstacles; (c) Comparison result of training time for different methods.}
\label{fig:comparison}
\vspace{-3mm}
\end{figure*}

To underscore the benefits of our methodology, we conducted comparative analyses with five SOTA baselines: PPO \citep{schulman2017proximal}, A2C \citep{mnih2016asynchronous}, DDPG \citep{lillicrap2015continuous}, SAC \citep{haarnoja2018soft}, and TD3 \citep{dankwa2019twin}, ensuring uniformity in neural network architectures, parameters, state-action representations, and reward functions across all benchmarks. Fig.~\ref{fig:comparison} highlights the superiority of our approach via enhanced performance metrics. Notably, the learning curve in (a) showcases our method's distinct advantage, while (b) reveals its significantly higher average success rate, consistently surpassing 90\% even with strict thresholds below 10 mm, confirming its precision and efficiency in trajectory tracking. Additionally, the training time comparison in (c) illustrates our method's accelerated convergence rate due to the integrated enhancements, demonstrating that AE and PF not only expedite policy improvement but also achieve higher rewards more rapidly. Our approach demonstrates efficiency by reaching fast convergence within 1.5 million timesteps, outperforming other methods in similar hardware and task conditions due to its time efficiency.

\subsubsection{Generalization Experiments}

\begin{figure}[!t]
    \includegraphics[width=1.0\linewidth]{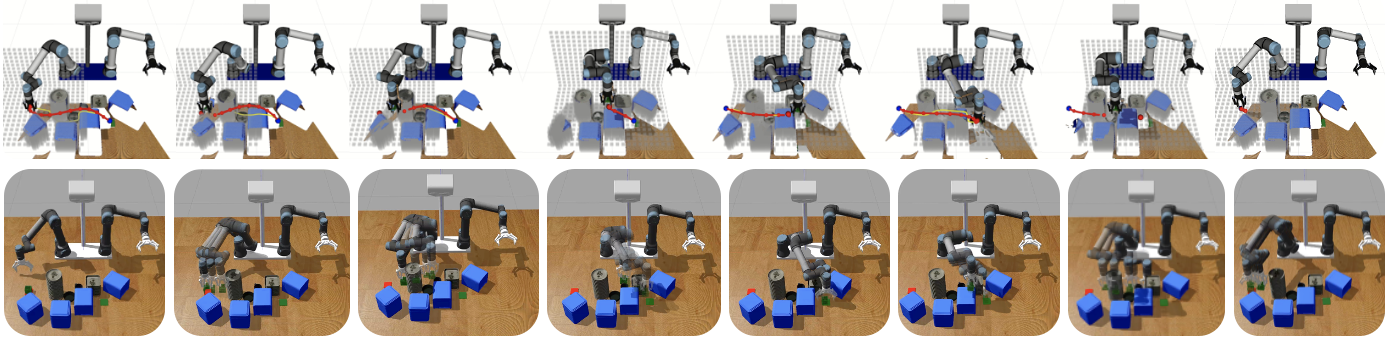}
\vspace{-5mm}
\caption{\textbf{Trajectory Planning in Gazebo:} The top image depicts the path planning process, where grey blocks represent detected obstacles and the red line traces the end-effector's path. The blue sphere marks the target, while the red sphere indicates the manipulator's current position. The green cube highlights the target location.}
\vspace{-3mm}
\label{fig:gazebo}
\end{figure}

\begin{figure}[!t]
    \includegraphics[width=1.0\linewidth]{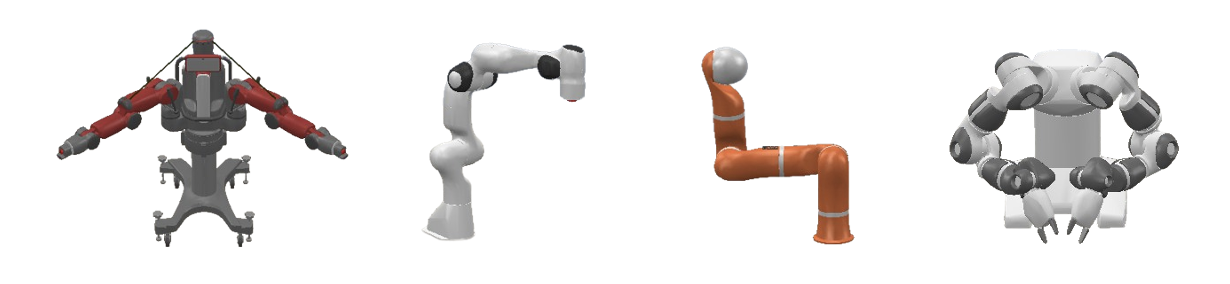}
    \par\noindent\makebox[0.24\linewidth][c]{\footnotesize(a) Baxter} 
    \makebox[0.24\linewidth][c]{\footnotesize(b) Franka Emika Panda} 
    \makebox[0.24\linewidth][c]{\footnotesize(c) Kuka LBR4+} 
    \makebox[0.24\linewidth][c]{\footnotesize(d) ABB Yumi} 
\caption{Robotic platforms used for evaluating the generalization of the proposed method. The experiments demonstrate our method's ability to perform effectively across a variety of robotic systems.}
\vspace{-3mm}
\label{fig:generalization}
\end{figure}

We assessed the robustness of our RL model in the Gazebo simulation environment as shown in Fig.~\ref{fig:gazebo}, where it consistently demonstrated reliable task performance, as highlighted in the accompanying video. Furthermore, to evaluate the efficiency of our proposed model across manipulators with varying DoF, we performed experiments on several robotic platforms, as shown in Fig.~\ref{fig:generalization}. The results indicate that our method performs comparably well on manipulators such as the Baxter, Franka Emika Panda, Kuka LBR4+, and ABB Yumi, confirming its ability to generalize effectively across a wide range of robotic systems. Details are available in supplementary videos.

\subsection{Implementation Evaluation in Motion Planning Tasks}
\subsubsection{Comparison Experiments}
\begin{figure*}[!t]
    \centering
    \includegraphics[width=1.0\linewidth]{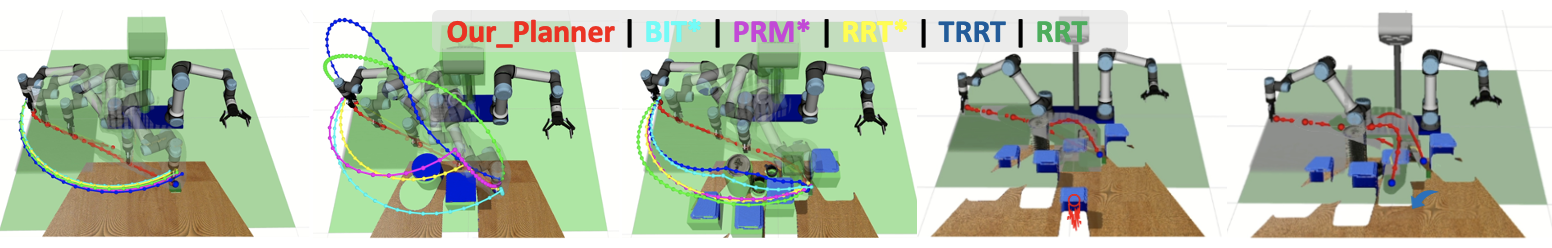}
    \par\noindent\makebox[0.18\linewidth][c]{\footnotesize(a) Task1} 
    \makebox[0.18\linewidth][c]{\footnotesize(b) Task2} 
    \makebox[0.18\linewidth][c]{\footnotesize(c) Task3} 
    \makebox[0.18\linewidth][c]{\footnotesize(d) Task4} 
    \makebox[0.18\linewidth][c]{\footnotesize(e) Task5} 
    \vspace{-3mm}
      \caption{Planner Comparison: Results from a single run for each planner across five tasks: \textbf{\textit{Task 1}} in an obstacle-free area; \textbf{\textit{Task 2}} with two additional obstacles; \textbf{\textit{Task 3}} in a cluttered scene; \textbf{\textit{Task 4}} featuring a moving obstacle near the goal; and \textbf{\textit{Task 5}} involving changing the goal as the end-effector nears it.}
      \label{fig:planner}
\end{figure*}

We evaluated our framework against five prevalent planners\citep{johnson2023learning} (RRT, TRRT, RRT*, PRM*, BIT*) using the Open Motion Planning Library (OMPL)\citep{sucan2012open}, as depicted in Fig.~\ref{fig:planner}. The evaluation encompassed five tasks, considering the robot's head and body as obstacles: \textbf{\textit{Task 1}} in an obstacle-free area; \textbf{\textit{Task 2}} with two additional obstacles; \textbf{\textit{Task 3}} in a cluttered space; \textbf{\textit{Task 4}} featuring a moving obstacle near the goal; \textbf{\textit{Task 5}} involves changing the goal as the end-effector nears it. We assessed planning performance with metrics: Length, Time, and Success Rate, and calculated the average values for each task and planner based on 100 trials. It's important to note that in tasks 1-3, all planners can successfully plan. However, as the difficulty increases, tasks 4 and 5 involve dynamic environments where obstacles move or targets change suddenly. In such cases, other planners struggle to adapt within the OMPL library, whereas our planner is still able to complete these tasks.

\begin{table*}[!t]
\centering
\caption{\fontsize{8pt}{8pt} \selectfont Result Comparisons of the Motion Planning Experiments Using Different Planners}
\label{table:time}
\fontsize{6pt}{6pt}\selectfont  
\begin{tabularx}{\textwidth}{X*{18}{>{\centering\arraybackslash}X}}
\toprule
 & \multicolumn{6}{c}{Length (m)} & \multicolumn{6}{c}{Time (sec)} & \multicolumn{6}{c}{Success Rate (\%)} \\
\cmidrule(lr){2-7} \cmidrule(lr){8-13} \cmidrule(lr){14-19}
& RRT & TRRT & RRT* & PRM* & BIT* & Ours & RRT & TRRT & RRT* & PRM* & BIT* & Ours & RRT & TRRT & RRT* & PRM* & BIT* & Ours \\ 
\midrule
Task1 & 2.04 & 2.03 & 1.88 & 1.88 & 1.99 & \textcolor{red}{\textbf{1.52}} & 0.19 & 0.24 & 5.04 & 5.05 & 2.92 & \textcolor{red}{\textbf{0.002}} & 100\% & 100\% & 100\% & 100\% & 100\% & \textcolor{red}{\textbf{100\%}} \\ 
Task2 & 2.20 & 2.13 & 2.02 & 2.13 & 2.11 & \textcolor{red}{\textbf{1.57}} & 0.27 & 0.43 & 5.06 & 5.08 & 4.19 & \textcolor{red}{\textbf{0.004}} & 99\% & 96\% & 99\% & 98\% & 100\% & \textcolor{red}{\textbf{100\%}} \\
Task3 & 2.33 & 2.22 & 2.12 & 2.06 & 1.96 & \textcolor{red}{\textbf{1.69}} & 0.79 & 1.04 & 5.06 & 5.07 & 5.08 & \textcolor{red}{\textbf{0.012}} & 99\% & 98\% & 98\% & 99\% & 100\% & \textcolor{red}{\textbf{100\%}} \\
Task4 & \ding{55} & \ding{55} & \ding{55} & \ding{55} & \ding{55} & \textcolor{red}{\textbf{1.72}} & \ding{55} & \ding{55} & \ding{55} & \ding{55} & \ding{55} & \textcolor{red}{\textbf{0.022}} & \ding{55} & \ding{55} & \ding{55} & \ding{55} & \ding{55} & \textcolor{red}{\textbf{100\%}} \\
Task5 & \ding{55} & \ding{55} & \ding{55} & \ding{55} & \ding{55} & \textcolor{red}{\textbf{1.76}} & \ding{55} & \ding{55} & \ding{55} & \ding{55} & \ding{55} & \textcolor{red}{\textbf{0.013}} & \ding{55} & \ding{55} & \ding{55} & \ding{55} & \ding{55} & \textcolor{red}{\textbf{100\%}} \\
\bottomrule
\end{tabularx}
\label{table:1}
\vspace{-3mm}
\end{table*}

Fig.~\ref{fig:planner} shows task comparisons. In static environments (tasks 1-3), our planner consistently achieves the shortest end-effector paths from start to goal. Details on manipulator states at each waypoint are available in supplementary videos. The planner maintains gradual end-effector orientation changes, beneficial for tasks involving grasping and human-like motions. In dynamic environments where traditional planners falter, our planner still efficiently plans trajectories in real-time. Table.~\ref{table:1} reveals our method's superiority in finding the quickest, shortest trajectories in static environments compared to others. It efficiently integrates 3D environmental data, such as point clouds, to outline sampling areas and adapts to dynamic scenarios with moving obstacles or targets.

\subsubsection{Motion Planning in Real World}
\begin{figure}[!t]
    \includegraphics[width=1.0\linewidth]{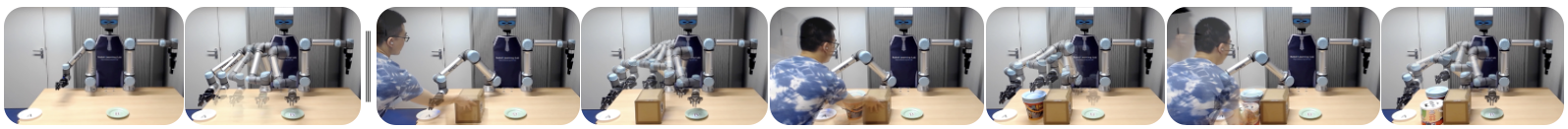}
    \includegraphics[width=1.0\linewidth]{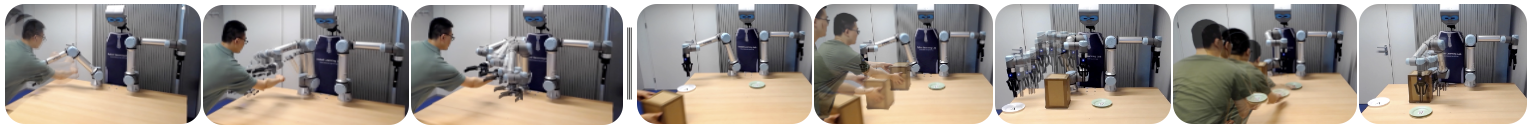}

\vspace{-5mm}
\caption{Motion planning experiments in the real world: The \textit{top-row}, from left to right, shows reaching from A (white plate) to B (green plate), obstacle addition during movement. The \textit{lower-row} shows reacting to an approaching person, the moving obstacle and changing the goal pose during execution.}
\vspace{-3mm}
\label{fig:mp_real}
\end{figure}

We integrated the proposed method into our robot for motion planning. Fig.~\ref{fig:mp_real} showcases its navigation capabilities in complex environments. The top row (left) illustrates a task with no obstacles, where our method moves directly from A to B in the shortest possible distance. The middle image shows a task where obstacles increase in real-time as the manipulator moves, but our planner ensures a safe path. The right image demonstrates a reactive task where the planner quickly responds to a person approaching the manipulator, highlighting its real-time planning capability. The bottom row depicts a scenario where an obstacle suddenly appears, and as the end-effector nears the goal, the goal changes. Despite these challenges, our planner plans successfully. Each experiment was conducted over five times, and the planner consistently executed successfully, given stable perception. Further details are provided in an accompanying video.

\subsection{Failure Cases}
In our real-robot experiments, we observed several distinct failure modes, which can be categorized into three primary types:

\textbf{Perception Failures due to Self-Occlusion:} During dynamic tasks, the manipulator itself sometimes occludes key parts of the scene. This issue arises because our Kinect sensor is mounted at a fixed head position, leading to partial or total blockage of the scene when the arm moves. This occlusion can degrade the quality of the environment map and affect motion planning. \textit{Possible solutions} include integrating multi-view or external camera setups (e.g., shoulder- or wrist-mounted RGB-D sensors) to reduce blind spots; implementing occlusion-aware perception pipelines that exclude the manipulator from segmentation or maintain a dynamic visibility map; and using partial map updates or temporal smoothing to retain information about occluded areas.

\textbf{Reactive Planning Limitations with Moving Obstacles:} A second category of failure occurs when the end-effector’s current pose becomes too close to a moving obstacle before the system has time to respond. This is particularly problematic when the obstacle's behavior is delayed or not predictable, resulting in invalid or unsafe trajectories. \textit{Potential solutions:} Potential strategies include incorporating predictive models of obstacle motion to anticipate movements and re-plan in advance; increasing the replanning frequency or employing hierarchical planners capable of responding more quickly at different levels (e.g., reactive control for short-term avoidance); and including safety buffers around moving entities during path planning to ensure proactive clearance.

\textbf{Goal-Change Failures due to Spatial-Temporal Constraints:} The third failure type arises during the goal-changing task. When the manipulator is already near the original goal and a sudden change occurs, there is often insufficient time to re-plan a trajectory before execution completes. This reflects the lag between perception, planning, and actuation. \textit{Potential solutions:} Potential strategies include incorporating fast-switchable motion primitives that can adapt execution mid-flight, maintaining low-latency planning pipelines or fallback reactive policies for nearby goal shifts, and introducing uncertainty-aware planning methods that anticipate possible changes and prepare contingencies.


\section{Discussion}
\label{sec:discussion}

Designing a specialized trajectory planner for manipulators capable of executing complex real-world tasks remains a significant challenge. In this section, we analyze how the proposed framework addresses these challenges and discuss our findings. Firstly, our perception-aware, B-spline-optimized kinodynamic planner provides time-parameterized trajectories that respect kinematic and dynamic constraints while enabling on-the-fly replanning. Empirically, Sec~\ref{sec:experiments} shows competitive path length and planning time in static scenes, and successful execution in dynamic scenes where classical sampling planners struggle. The waypoint interface further standardizes the handoff from task space to the joint-space controller, easing deployment across different arms. Secondly, the proposed RL controller directly maps task-space waypoints to safe joint commands without requiring explicit inverse kinematics, which is advantageous near kinematic singularities or in clutter. In aggregate, the controller sustains accurate reaching while maintaining link safety, as evidenced by high success rates under tight thresholds. Finally, the AE component smooths action sampling and reduces variance, accelerating convergence without suppressing exploration. The PF mechanism introduces confidence-aware critic updates that better align actor and value learning. Ablations show consistent gains from each module; in combination, AE+PF achieves higher cumulative reward, faster training, and more stable performance in dynamic or partially observable conditions.

In summary, the proposed planner–controller stack provides a fast, manipulator-specific trajectory generation pipeline coupled with a reinforcement learning controller tailored for precise, collision-aware reaching. By stabilizing PPO through the integration of AE and PF, the framework achieves improved efficiency, robustness, and adaptability for real-time manipulation tasks.

\section{Conclusions}
\label{sec:conclusion}
This paper presents a fast trajectory planning system for a 6-DoF manipulator. The fast trajectory planner integrates vision-based trajectory planning in task space with a controller for joint-level obstacle avoidance, based on reinforcement learning. For trajectory planning, we leverage FSA-augmented 3D mapping to enhance environmental perception and integrate kinodynamic path searching with B-spline-based optimization, ensuring the generated trajectories are both efficient and safe. In the case of trajectory tracking, we propose an innovative RL-based controller that incorporates two significant advancements within the PPO framework. First, we rigorously evaluate the implementation of action ensembles across diverse distributions through comprehensive ablation studies and comparative analysis. Second, the integration of policy feedback improves reward acquisition and accelerates convergence, effectively shortening the training time. 

To validate the effectiveness of our method, we conduct extensive sets of simulations and real-world experiments. We compare our method with five other planners, demonstrating that our method consistently achieved the shortest trajectories at the fastest speeds in static environments. Additionally, our planner performs effectively in dynamic and complex tasks, dynamically navigating through obstacles in various environments. In the future, we plan to explore learning-based replanning strategies to replace heuristic triggers, enabling more adaptive and efficient responses in dynamic environments. Incorporating lightweight model-based components into our model-free framework may enhance data efficiency and improve the feasibility of generated motions. Furthermore, integrating our approach with high-level task and motion planning (TAMP) frameworks could support more complex, sequential behaviors such as grasp–place–regrasp in real-world scenarios.


\appendix
\section{Appendix Section}
\label{appendix}

This appendix primarily elaborates on the details of the paper. Initially, it delves into the kinematic model of the robot and computation of errors in state representation within the enhanced Proximal Policy Optimization (PPO) algorithm. Subsequently, it explains the methodology for calculating distances between links and obstacles. It then presents the training workspace for a 6-degree-of-freedom (DoF) manipulator across all evaluated algorithms. Furthermore, results from ablation studies on Action Ensembles with various distributions are displayed, illustrating the impact of different hyperparameters on performance for each distribution.

\subsection{Kinematic Model}

The experimental platform in this study is the Universal Robots UR5e, an anthropomorphic manipulator widely used in industry. The UR5e features 6 DoF, a maximum payload capacity of $5~\text{kg}$, and a reach of $850~\text{mm}$, allowing accurate positioning of the end-effector in arbitrary poses within its workspace. In our framework, the RL policy is designed to output the desired joint positions for all $n$ joints directly, such that the action at time step $t$ is represented as $\boldsymbol{a}_t = \boldsymbol{q}_t \in \mathbb{R}^n$. This design choice removes the need for intermediate inverse kinematics (IK) solvers or velocity-based differential control during deployment, streamlining the control pipeline. The corresponding end-effector pose, encompassing both position and orientation, is then obtained through the conventional forward kinematics (FK) formulation \citep{diaz2025neural}.

\begin{table}[h]
\centering
\caption{UR5e DH Parameters (radians)}
\begin{tabular}{c c c c c}
\hline
$i$ & $\alpha_{i-1}$ & $a_{i-1}$ [m] & $d_i$ [m] & $\theta_i$ [rad] \\
\hline
1 &$ \tfrac{\pi}{2}$   & 0        & 0.1625 & $(-2\pi,\,2\pi)$ \\
2 & 0                & -0.425   & 0      & $(-2\pi,\,2\pi)$ \\
3 & 0                & -0.3922  & 0      & $(-2\pi,\,2\pi)$ \\
4 & $\tfrac{\pi}{2}$   & 0        & 0.1333 & $(-2\pi,\,2\pi)$ \\
5 & $-\tfrac{\pi}{2}$  & 0        & 0.0997 & $(-2\pi,\,2\pi)$ \\
6 & 0                & 0        & 0.0996 & $(-2\pi,\,2\pi)$ \\
\hline
\end{tabular}
\label{tab:dh}
\end{table}

The FK defines the mapping from the robot's joint configuration to the pose of its end-effector. In this work, the FK model of the UR5e manipulator is formulated using the standard Denavit–Hartenberg (D–H) convention, with the parameters listed in Table~\ref{tab:dh}. These parameters specify the relative position and orientation between consecutive links in the kinematic chain. The transformation from the $(i-1)$-th link frame to the $i$-th link frame is expressed as:

\begin{equation}
{}^{i-1}_{i}T = \mathrm{Trot}_z(\theta_i)\,\mathrm{Transl}_z(d_i)\,\mathrm{Transl}_x(a_i)\,\mathrm{Trot}_x(\alpha_i),
\label{eq:DH}
\end{equation}
where $\mathrm{Trot}$ and $\mathrm{Transl}$ denote rotation and translation operations, respectively, around the indicated axes. The complete FK from the robot base frame $b$ to the end-effector frame $e$ is obtained by chaining these transformations:
\begin{equation}
{}^{b}_{e}T = \prod_{i=1}^{6} {}^{i-1}_{i}T(a_i, \alpha_i, d_i, \theta_i),
\label{eq:FK}
\end{equation}
where $\theta_r(t) = [\theta_1(t), \ldots, \theta_6(t)]$ is the robot's joint configuration vector. The UR5e features 6 revolute joints, $\mathbf{q} = (q_1, q_2, q_3, q_4, q_5, q_6)$. These parameters, combined with the D–H convention, allow the analytical computation of the end-effector pose. Formally, FK is expressed as:

\begin{equation}
\mathbf{p} = f(\mathbf{q}),
\label{eq:kinematics}
\end{equation}
where $f: \mathbb{R}^n \to \mathbb{R}^m$ maps the $n$ joint coordinates to an $m$-dimensional task-space pose vector, typically comprising both Cartesian position and orientation.

In our framework, FK serves as a critical interface between joint-space control and Cartesian-space task objectives, fulfilling three primary roles:
\begin{itemize}
    \item \textbf{State representation:} Computing the current end-effector pose for inclusion in the observation vector.
    \item \textbf{Reward shaping:} Evaluating the pose error with respect to the target and incorporating it into the reward function.
    \item \textbf{Task evaluation:} Verifying during testing whether the final pose meets the target within a specified tolerance.
\end{itemize}

\subsection{Pose Error Calculation}
The state is represented as:
\begin{equation}
    \bm s_t = <\bm q_t, \bm p_e, \bm p_t, \bm {error}, \bm d_{obs}>
\end{equation}
The \(\bm{error}\) vector measures the difference between the end-effector and the target, integrating the sum of distance errors from three points defined along orthogonal axes and the orientation error. The computation proceeds as follows:

\subsubsection{Generating Tri-Point Representations for Target and End-Effector Along Orthogonal Axes}
To describe the process mathematically for generating three points on both the target and the end-effector. As depicted in Fig.~\ref{fig:1}, the following steps are outlined:

\paragraph{1. Convert Quaternion to Euler Angles} Convert the orientation of both the target and the end-effector from quaternions to Euler angles (roll \( \phi \), pitch \( \theta \), and yaw \( \psi \)):

\begin{equation}
\text{Quaternion} \rightarrow (\phi, \theta, \psi)
\end{equation}

\paragraph{2. Calculate Rotation Matrices} Construct the following rotation matrices for roll, pitch, and yaw based on Euler angles:

\emph{Roll rotation matrix} \( R_{\phi} \):

\begin{equation}
R_{\phi} = \begin{bmatrix} 
1 & 0 & 0 \\ 
0 & \cos(\phi) & -\sin(\phi) \\ 
0 & \sin(\phi) & \cos(\phi) 
\end{bmatrix}
\end{equation}

\emph{Pitch rotation matrix} \( R_{\theta} \):

\begin{equation}
R_{\theta} = \begin{bmatrix} 
\cos(\theta) & 0 & \sin(\theta) \\ 
0 & 1 & 0 \\ 
-\sin(\theta) & 0 & \cos(\theta) 
\end{bmatrix}
\end{equation}

\emph{Yaw rotation matrix} \( R_{\psi} \):

\begin{equation}
R_{\psi} = \begin{bmatrix} 
\cos(\psi) & -\sin(\psi) & 0 \\ 
\sin(\psi) & \cos(\psi) & 0 \\ 
0 & 0 & 1 
\end{bmatrix}
\end{equation}

\paragraph{3. Generate Points} For each rotation (roll, pitch, and yaw), calculate a point along the respective axis at a distance \( d \) from the position \( \textbf{p} \) (target or end-effector position):

\begin{equation}
\textbf{p}_{\phi} = \textbf{p} + d \cdot R_{\phi}[:,0]
\end{equation}

\begin{equation}
\textbf{p}_{\theta} = \textbf{p} + d \cdot R_{\theta}[:,1]
\end{equation}

\begin{equation}
\textbf{p}_{\psi} = \textbf{p} + d \cdot R_{\psi}[:,2]
\end{equation}

\subsubsection{Computing Distances Between Corresponding Points}
Then calculate the distance between the corresponding points:

\paragraph{1. Calculate Distances} Compute the Euclidean distance between corresponding points generated for the target and the end-effector:

Distance between points along roll axis \( D_{\phi} \), pitch axis \( D_{\theta} \), and yaw axis \( D_{\psi} \):

\begin{equation}
D_{\phi} = \| \textbf{p}_{\phi,\text{end\_effector}} - \textbf{p}_{\phi,\text{target}} \|
\end{equation}

\begin{equation}
D_{\theta} = \| \textbf{p}_{\theta,\text{end\_effector}} - \textbf{p}_{\theta,\text{target}} \|
\end{equation}

\begin{equation}
D_{\psi} = \| \textbf{p}_{\psi,\text{end\_effector}} - \textbf{p}_{\psi,\text{target}} \|
\end{equation}

\noindent Here, \( \textbf{p} \) represents the position vector, \( d \) is the predefined distance from the position to the point, and \( R_{\phi}[:,0] \), \( R_{\theta}[:,1] \), \( R_{\psi}[:,2] \) represent the column vectors of the rotation matrices corresponding to the roll, pitch, and yaw axes, respectively. The distances \( D_{\phi} \), \( D_{\theta} \), and \( D_{\psi} \) quantify the spatial discrepancy between the target and end-effector orientations and positions in three-dimensional space.

\subsubsection{Calculating the Shortest Angular Difference Between Quaternions for Target and End-Effector}

To steer the learning strategy towards achieving the target orientation, the angular difference between the target and the end-effector is incorporated. The formula for calculating the shortest angular difference between two quaternions, \( \textbf{quat1} \) and \( \textbf{quat2} \), is mathematically delineated as follows:

\paragraph{1. Calculate the dot product of \( \textbf{quat1} \) and \( \textbf{quat2} \)}

\begin{equation}
\text{dot\_product} = \textbf{quat1} \cdot \textbf{quat2}
\end{equation}

\paragraph{2. Ensure the dot product is within the range \lbrack-1, 1\rbrack to account for floating-point inaccuracies}

\begin{equation}
\text{dot\_product} = \text{clip}(\text{dot\_product}, -1.0, 1.0)
\end{equation}

\paragraph{3. Determine the angle difference}

\begin{equation}
\text{angle\_diff} = 2 \cdot \arccos(\text{dot\_product})
\end{equation}

\paragraph{4. Compute the shortest angular difference \(\Delta \theta\) between two angles, \( \theta_1 = 0 \) and \( \theta_2 = \text{angle\_diff} \), while ensuring the result is within the range \(\lbrack 0, \pi \rbrack\)}

\begin{equation}
\Delta \theta = \min( |\theta_2 - \theta_1|, 2\pi - |\theta_2 - \theta_1|)
\end{equation}

\begin{equation}
\Delta \theta = |\theta_2 - \theta_1|
\end{equation}

\begin{equation}
\text{If } \Delta \theta > \pi, \text{ then } \Delta \theta = 2\pi - \Delta \theta
\end{equation}

Thus, the expression for calculating the shortest angular difference in radians between two quaternions is:

\begin{equation}
\Delta \theta = \left\{ \begin{array}{ll} 
|\theta_2 - \theta_1| & \mbox{if } |\theta_2 - \theta_1| \leq \pi \\ 
2\pi - |\theta_2 - \theta_1| & \mbox{if } |\theta_2 - \theta_1| > \pi 
\end{array} \right.
\end{equation}

\noindent where \( \theta_2 = 2 \cdot \arccos(\text{dot\_product}) \) and \( \theta_1 = 0 \).

Ultimately, the error vector, $\bm{error}$, is computed as $\bm{error} = (D_{\phi} + D_{\theta} + D_{\psi}, \Delta \theta)$, encapsulating both the cumulative distance discrepancies along the orthogonal axes and the angular difference.

\begin{figure}[!t]
      \centering
      \includegraphics[width=0.35\linewidth]{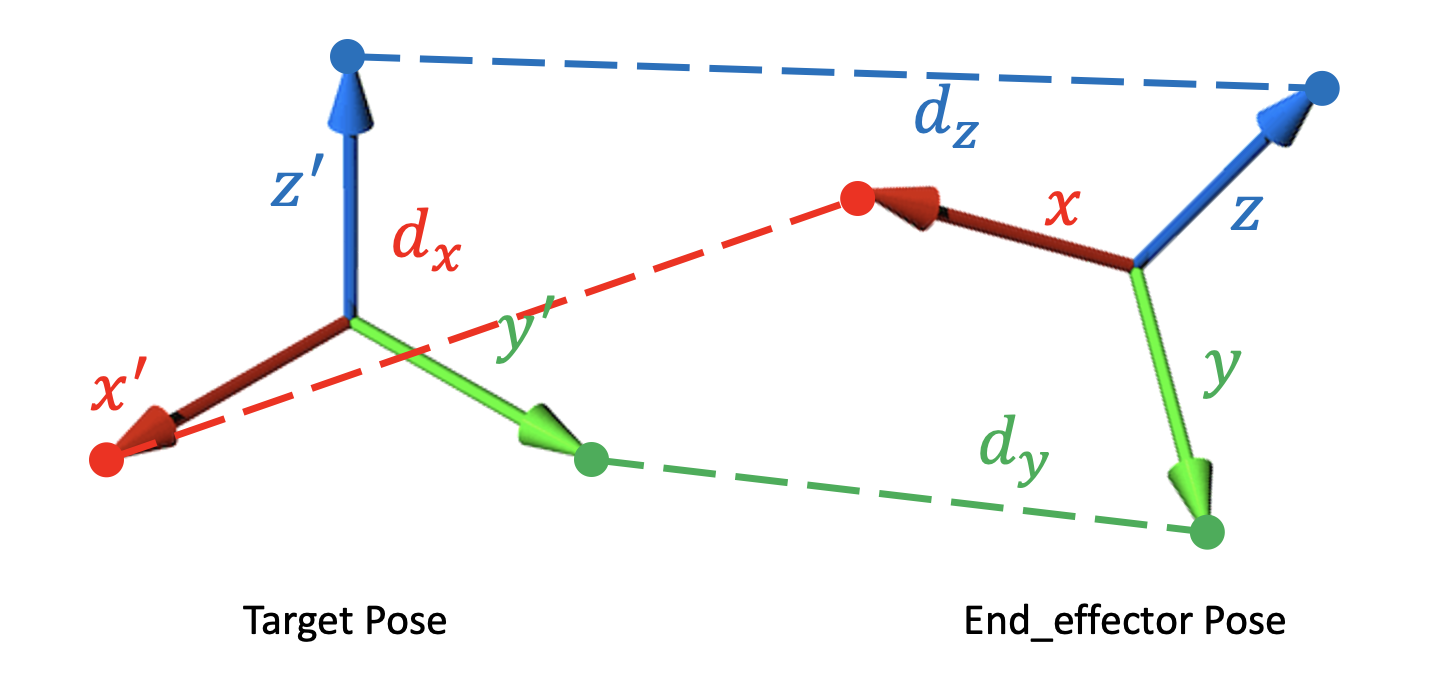}
      \includegraphics[width=0.35\linewidth]{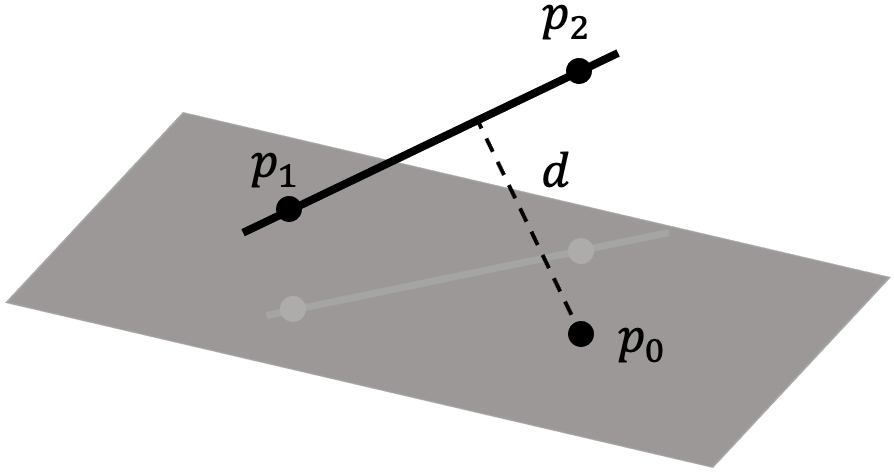}
      \includegraphics[width=0.23\linewidth]{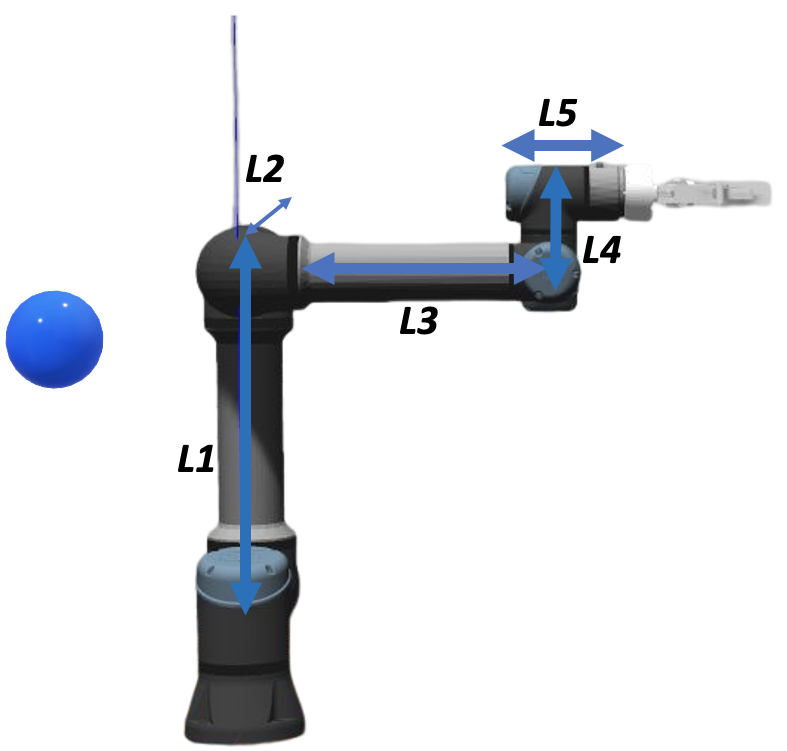}
      \par\noindent\makebox[0.35\linewidth][c]{\footnotesize(a)} 
      \makebox[0.35\linewidth][c]{\footnotesize(b)} 
      \makebox[0.23\linewidth][c]{\footnotesize(c)} 
      \caption{(a) The difference calculation between the end-effector and the target, integrating the sum of distance errors from three points defined along orthogonal axes. (b) Distance between a point and a space line. (c) \(\bm d_{obs}\) is the obstacle (blue) to 5-link distance. }
      \label{fig:1}
\end{figure}

\subsection{Distances between Links and Obstacles}
The $\bm d_{obs}$ in state representation (Equ.~\ref{equ:1}) is the shortest distance between obstacles and links in space. As depicted in Fig.~\ref{fig:1} (b) and (c), to calculate the distance between the obstacle and each link in joint space, we transform it into a geometric problem to find the shortest distance between any point in space and different links. We provide an example in Fig.~\ref{fig:1} (b) to make it clearer. Let a 3D line be specified by 2 points, $p_1 = (x_1, y_1, z_1)$ and $p_2 = (x_2, y_2, z_2)$, where $\cdot$ represents the dot product. Therefore, a vector along the line is given by the following equation:

\begin{equation}
    \bm v = \begin{bmatrix} 
        x_1 + (x_2 - x_1) t \\ 
        y_1 + (y_2 - y_1) t \\ 
        z_1 + (z_2 - z_1) t 
    \end{bmatrix}
\end{equation}

The squared distance between a point on the line with parameter $t$ and a point $p_0 = (x_0, y_0, z_0)$ is therefore:
\begin{equation}
   \begin{aligned}
    d^2 = [(x_1-x_0)+(x_2-x_1)t]^2\\
    +[(y_1-y_0)+(y_2-y_1)t]^2\\
    +[(z_1-z_0)+(z_2-z_1)t]^2
   \end{aligned}
   \label{equ:a20}
\end{equation}
Set $d(d^2)/dt=0$ and solve for $t$ to obtain the shortest distance:
\begin{equation}
    \bm t = -\dfrac{(x_1-x_0)\cdot(x_2-x_1)}{|x_2-x_1|^2}
    \label{equ:a21}
\end{equation} 
The shortest distance can then be calculated by plugging Eq.~\ref{equ:a21} back into Eq.~\ref{equ:a20}. Thus, as shown on the right side of Fig.~\ref{fig:1} (c), we consider each link of the robot as the line and the obstacle as the point. We then calculate the shortest distance between every obstacle and link. 


\subsection{Workspace}
\begin{figure*}[!t]
    \includegraphics[width=0.32\linewidth]{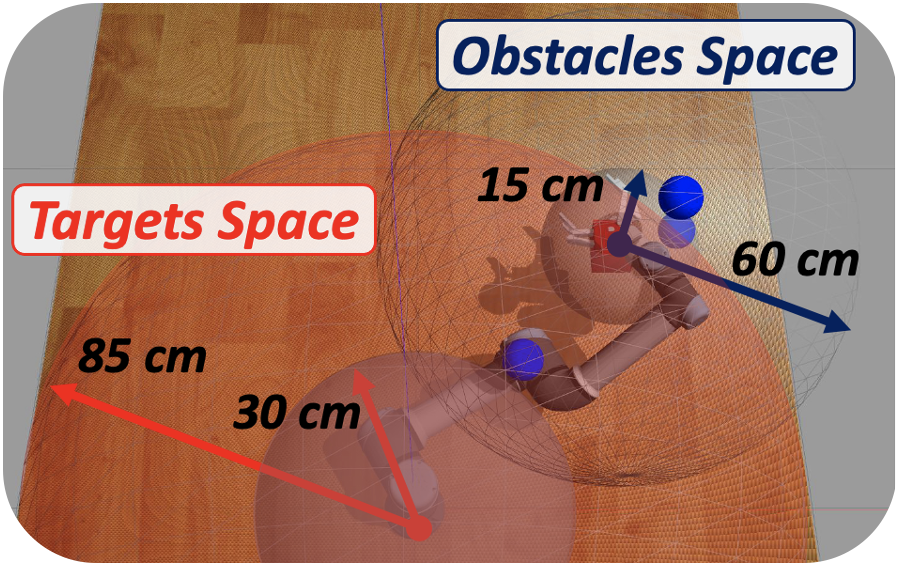}
    \includegraphics[width=0.32\linewidth]{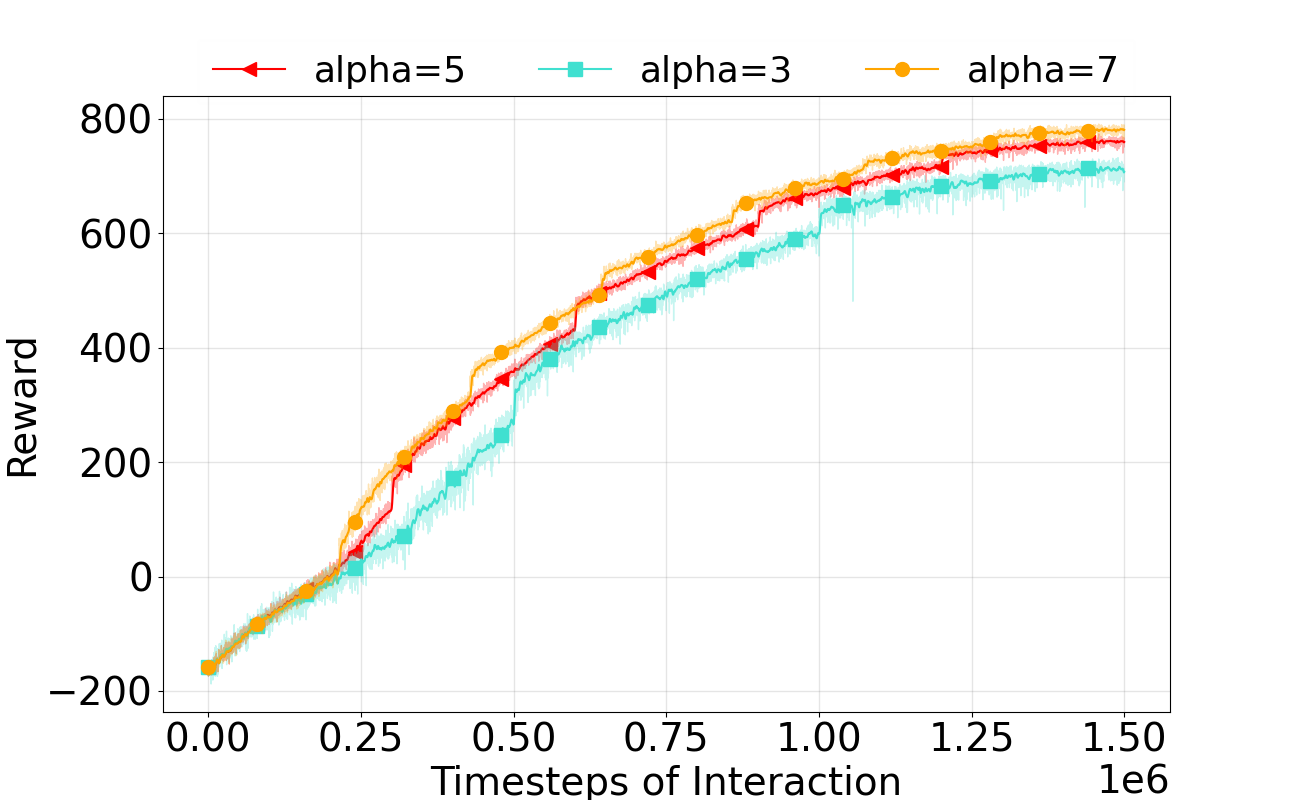}
    \includegraphics[width=0.32\linewidth]{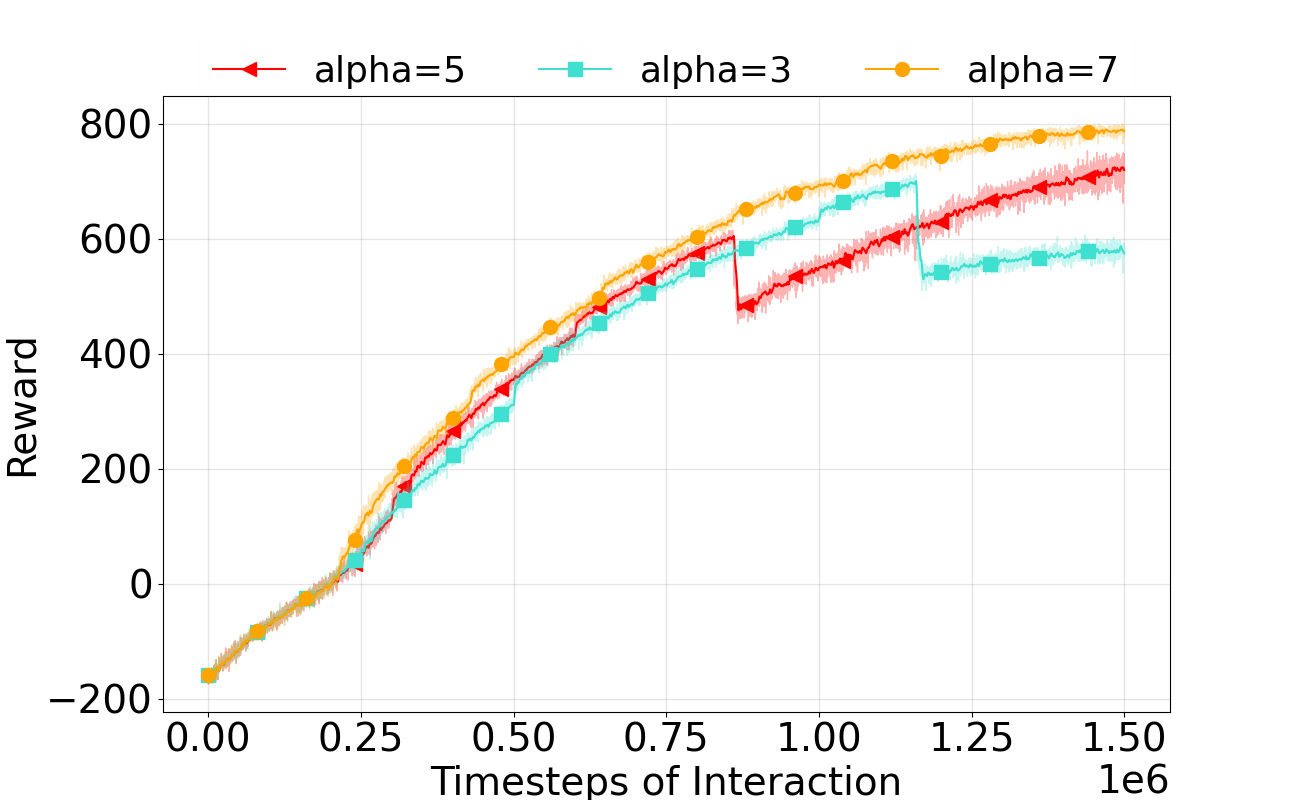}
    \par\noindent\makebox[0.32\linewidth][c]{\footnotesize(a) Workspace} 
    \makebox[0.32\linewidth][c]{\footnotesize(b) PPO\_PF\_AEL} 
    \makebox[0.32\linewidth][c]{\footnotesize(c) PPO\_PF\_AEP}
    \includegraphics[width=0.32\linewidth]{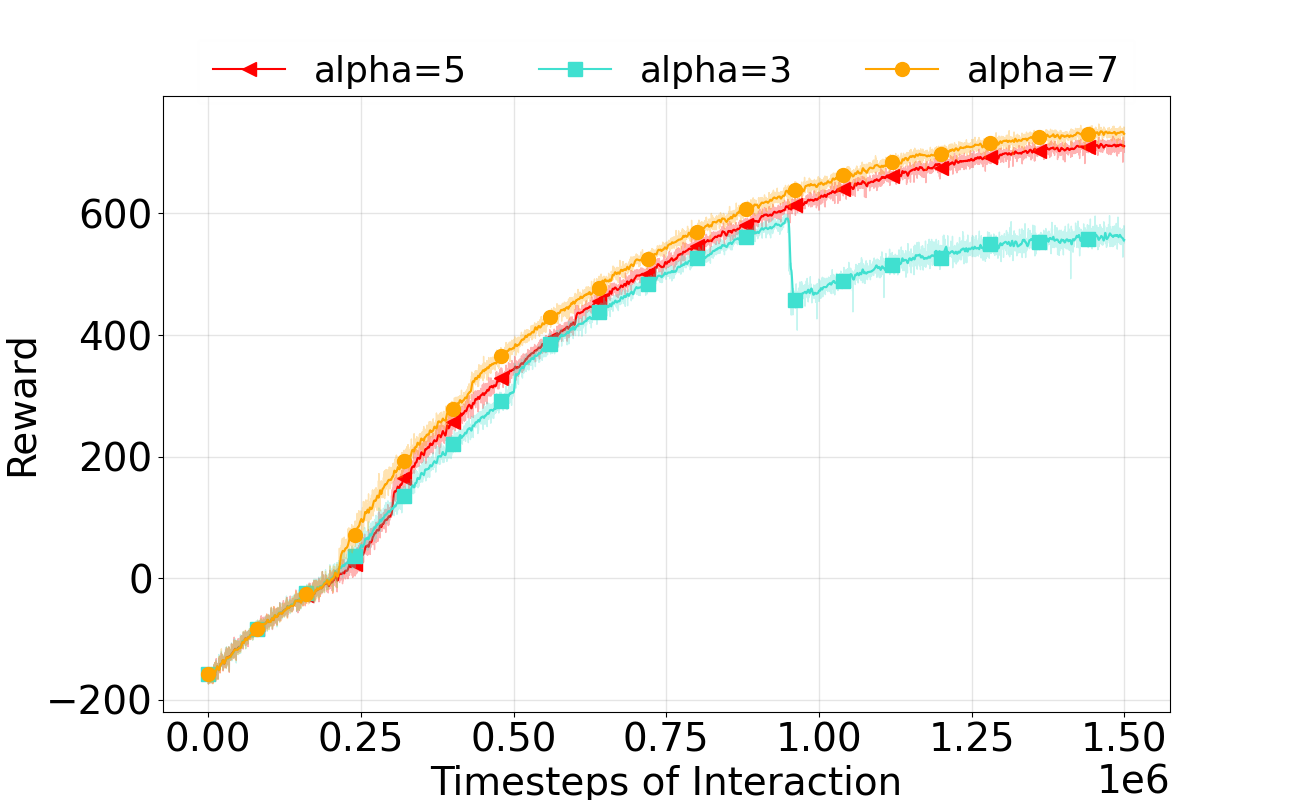}
    \includegraphics[width=0.32\linewidth]{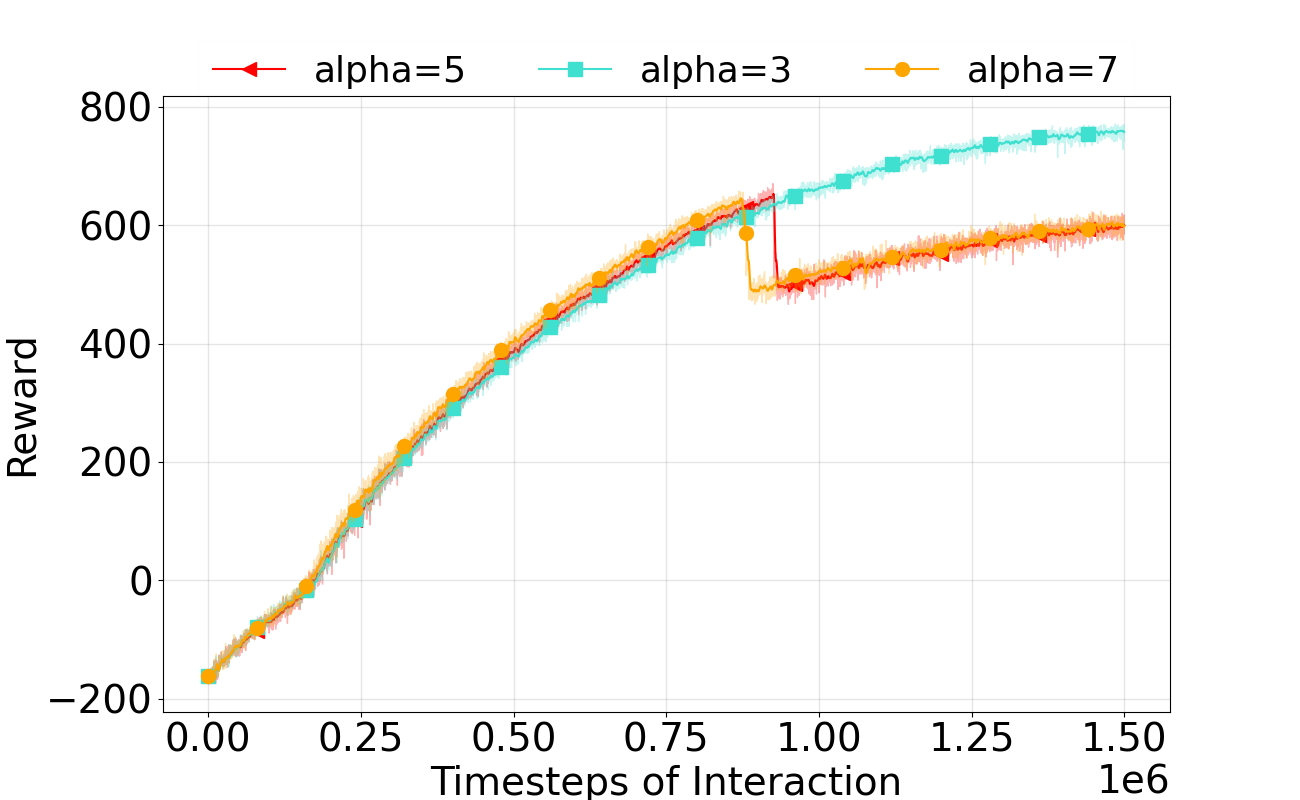}
    \includegraphics[width=0.32\linewidth]{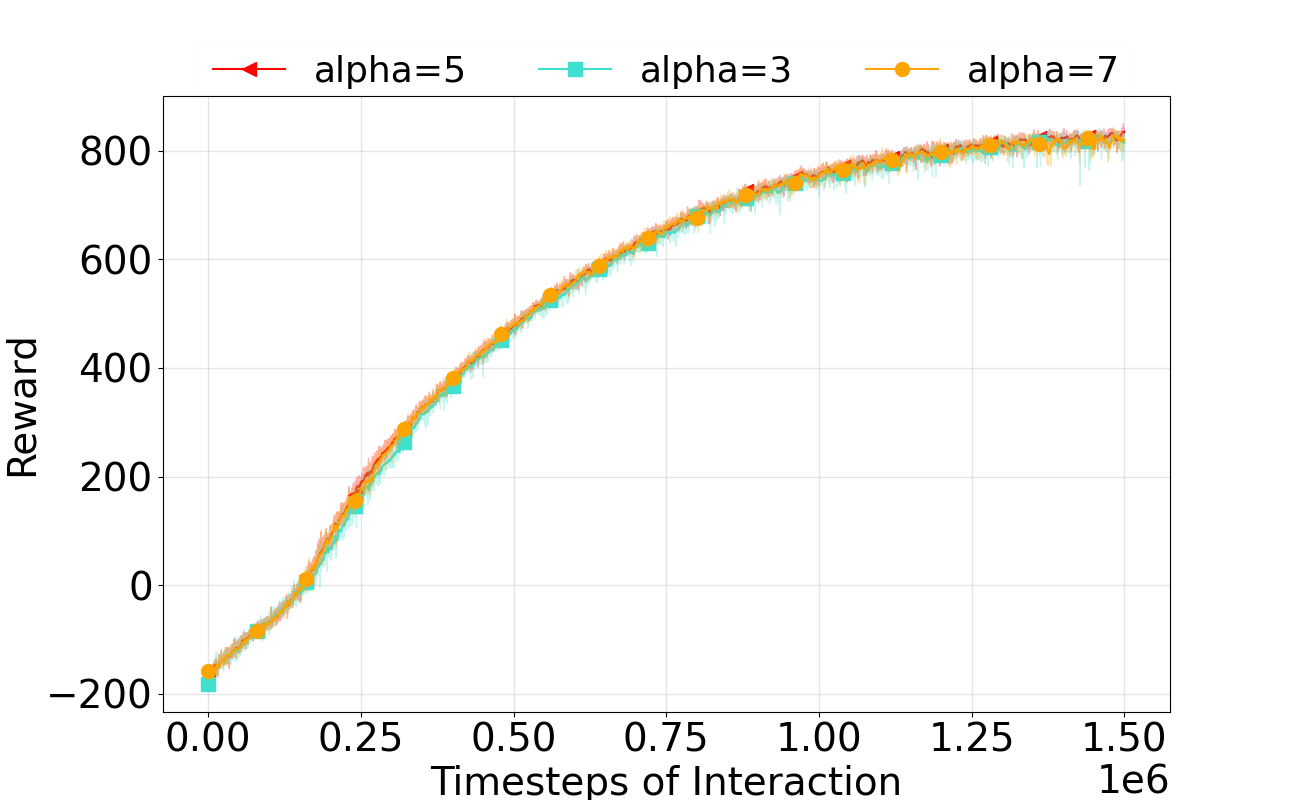}
    \par\noindent\makebox[0.32\linewidth][c]{\footnotesize(d) PPO\_PF\_AEB} 
    \makebox[0.32\linewidth][c]{\footnotesize(e) PPO\_PF\_AEE} 
    \makebox[0.32\linewidth][c]{\footnotesize(f) PPO\_PF\_AEW} 
\caption{Workspace and Summary of Ablation Studies with Varying $\alpha$: The red line represents $\alpha=3$, turquoise denotes $\alpha=5$, and orange signifies $\alpha=7$, across 5 random seeds.}
\label{fig:3}
\end{figure*}

As shown in Fig.~\ref{fig:3} (a), the experimental workspace takes the form of a sphere with a radius of \(85 \text{ cm}\) but excludes a \(30 \text{ cm}\) radius cylinder at the base of the robot. During training, targets are randomly located within the workspace, and three \(5 \text{ cm}\)-diameter spherical obstacles are positioned either near the target or the manipulator's links to represent regions to avoid in joint space. Obstacle spheres are randomly spawned within a quarter-spherical annulus, centered at the target, with the major and minor radii of \(60 \text{ cm}\) and \(15 \text{ cm}\), respectively.

\subsection{Action Ensembles Ablation Experiments}

\begin{itemize}
  \item AE with Linear (AEL): Linear improvement for sample selection, defined specifically as follows:

\begin{equation}
    i = 1+ \alpha \dfrac{e_n}{e_a}, \bm a_{t,j}\sim N(\mu_\theta(s_t), \delta_\theta), \bm a_t = \underset{j}{mean}(\bm a_{t,j})
\end{equation}
where $\alpha$ serves as the hyper-parameter influencing the rate of convergence. with $j$ varying from 1 up to but not including $i$. $e_n$ and $e_a$ represent the counts of the current and total episodes, respectively. As shown in Fig.~\ref{fig:3} (b), we tested three different $\alpha$. 

  \item AE with Poisson Distribution (AEP): Poisson distribution for sample selection, defined specifically as follows:

\begin{equation}
    i\sim clip(Poisson(\beta), 1, \beta),\quad \beta = 1+ \alpha \dfrac{e_n}{e_a}
\end{equation}
\begin{equation}
    \bm a_{t,j}\sim N(\mu_\theta(s_t), \delta_\theta),\quad \bm a_t = \underset{j}{mean}(\bm a_{t,j})
\end{equation}
where \(\beta\) represents the parameter of the Poisson distribution. As shown in Fig.~\ref{fig:3} (c), we tested three different $\alpha$. 

  \item AE with Beta Distribution (AEB): Beta distribution for sample selection, defined specifically as follows:

\begin{equation}
    i\sim clip(Beta(a, b), 1, a) 
\end{equation}
\begin{equation}
    a = 1+ \alpha \dfrac{e_n}{e_a} \quad b = 1+ \beta \dfrac{e_n}{e_a} 
\end{equation}
\begin{equation}
    \bm a_{t,j}\sim N(\mu_\theta(s_t), \delta_\theta),\quad \bm a_t = \underset{j}{mean}(\bm a_{t,j})
\end{equation}
where $a$ and $b$ represent the parameters of the Beta distribution. $\alpha$ and $\beta$ are the hyper-parameters influencing the convergence rate. As shown in Fig.~\ref{fig:3} (d), we tested three different $\alpha$. 
  
  \item AE with Exponential Distribution (AEE): Exponential distribution for sample selection, defined specifically as follows:

\begin{equation}
    i\sim clip(Exp(\lambda), 1, \lambda),\quad \lambda = 1/(1 + \alpha \dfrac{e_n}{e_a})
\end{equation}
\begin{equation}
    \bm a_{t,j}\sim N(\mu_\theta(s_t), \delta_\theta),\quad \bm a_t = \underset{j}{mean}(\bm a_{t,j})
\end{equation}
where $\lambda$ represents the parameter of the Exponential distribution, $\alpha$ is the hyper-parameters influencing the convergence rate. As shown in Fig.~\ref{fig:3} (e), we tested three different $\alpha$. 

  \item AE with Weibull Distribution (AEW): Weibull distribution for sample selection, defined specifically as follows:

\begin{equation}
    i\sim clip(Weibull(k, \lambda), 1, \lambda)
\end{equation}
\begin{equation}
    k = 1+ \alpha \dfrac{e_n}{e_a} \quad \lambda = 1+ \beta \dfrac{e_n}{e_a}
\end{equation}
\begin{equation}
    \bm a_{t,j}\sim N(\mu_\theta(s_t), \delta_\theta),\quad \bm a_t = \underset{j}{mean}(\bm a_{t,j})
\end{equation}
where $k$ and $\lambda$ represent the parameters of the Weibull distribution. $\alpha$ and $\beta$ are the hyperparameters influencing the convergence rate. As shown in Fig.~\ref{fig:3} (f), we tested three different $\alpha$. 

\end{itemize}

As illustrated in Fig.~\ref{fig:3} (b)-(f), the performance of AEW method appears largely consistent across different distribution parameter variations.

\textbf{Theoretical Comparison Across Distributions:} To further understand why the Weibull-based AE performs best among the proposed action ensemble variants, we provide a theoretical comparison based on their distributional properties and alignment with key demands of reinforcement learning, balancing exploration and exploitation, supporting adaptive learning schedules, and ensuring stable convergence.

\begin{itemize}
    \item \textbf{Linear (AEL):} Offers deterministic weighting growth based on estimated advantage. While simple and easy to tune, it lacks stochasticity and diversity in action sampling. This can lead to overconfidence in early-stage suboptimal policies and poor exploration in complex or noisy environments.

    \item \textbf{Poisson (AEP):} Models discrete count-based sampling around a rate parameter $\beta$. However, the distribution's support is non-negative integers with fixed mean-variance coupling $\text{Var}(X) = \mathbb{E}[X]$, making it relatively rigid. Its symmetry and limited skew control restrict nuanced adaptation of action weights during different learning phases.

    \item \textbf{Beta (AEB):} A flexible continuous distribution on $[0, 1]$ controlled by shape parameters $a$ and $b$. It offers a range of modes (e.g., U-shape, unimodal) depending on parameter values. However, it lacks an explicit exponential tail, which can be useful for controlling long-shot exploration, and requires careful parameter balancing to avoid biasing weight allocation prematurely.

    \item \textbf{Exponential (AEE):} A memoryless distribution defined by a single rate parameter $\lambda$, encouraging consistent exploration. While beneficial early on, its fixed shape (constant right-skew) cannot be adjusted to favor exploitation in later training stages, limiting convergence speed.

    \item \textbf{Weibull (AEW):} Extends the exponential by introducing a shape parameter $k$, enabling precise control over tail behavior and skewness:
    \begin{itemize}
        \item For $k < 1$, AEW produces heavy right-skew, favoring small weights for most actions and large weights for a few, enhancing early-stage exploration.
        \item For $k > 1$, the distribution becomes left-skewed, concentrating weights on high-advantage actions to promote exploitation and convergence.
        \item At $k = 1$, it recovers the exponential, naturally bridging both phases.
    \end{itemize}
    
    This adaptability allows AEW to transition from exploration to exploitation across training epochs smoothly. Additionally, Weibull’s continuous support and tunable scale parameter $\lambda$ allow better alignment with normalized advantage values. Its controlled tail behavior also mitigates instability from assigning excessively high weights to outlier actions.
\end{itemize}

Overall, AEW uniquely supports a dynamic action weighting scheme that evolves with the agent’s confidence, offering temporal adaptivity, distributional expressiveness, and stability, key traits not jointly present in other tested distributions.

Based on our theoretical understanding and empirical findings, we attribute the superior performance of AEW to the following reasons:

\begin{enumerate}
    \item \textbf{Adaptive Exploration Control:} The tunable shape parameter $k$ in the Weibull distribution enables AEW to \textit{modulate exploration intensity over time}. As mentioned above, $k < 1$ encourages broad exploration, while $k > 1$ supports focused convergence. This flexibility allows AEW to adapt naturally to different phases of training. Our learning curve in Fig.~\ref{fig:ablation} (a) and stability analysis in Fig.~\ref{fig:3} (f) illustrate this adaptive behavior.
    
    \item \textbf{Stability Across Hyperparameters:} Compared to other distributions like Beta or Poisson, AEW exhibits \textit{lower sensitivity to hyperparameter variations}, especially in $\alpha$. As shown in Fig.~\ref{fig:3} (f), performance remains consistent across different $\alpha$ values, making AEW more robust and practical for real-world systems where hyperparameter tuning is often expensive or impractical.
    
    \item \textbf{Smooth Uncertainty Reduction:} By using Weibull-guided sampling in the ensemble, AEW facilitates a \textit{gradual reduction of policy uncertainty}, avoiding abrupt changes in action selection. This leads to smoother learning curves and helps prevent convergence to suboptimal local minima, a critical feature for robotic control tasks that require precision and stability.
    
    \item \textbf{Empirical Advantage in Early and Mid Training Phases:} As demonstrated in our ablation study (Sec~\ref{sec:experiments}, Fig.~\ref{fig:ablation}), AEW leads to \textit{faster learning} in the early training phase and \textit{sustains high performance} throughout. This is particularly valuable in high-dimensional trajectory planning tasks where sample efficiency and early progress are essential.
\end{enumerate}

\section*{CRediT authorship contribution statement}
\textbf{Yongliang Wang:} Conceptualization, Methodology, Validation, Investigation, Software, Writing – original draft. \textbf{Hamidreza Kasaei:} Supervision, Software, Writing - review \& editing.

\section*{Declaration of competing interest}
The authors declare that they have no known competing financial inter-
ests or personal relationships that could have appeared to influence the work
reported in this paper.

\section*{Data availability}
Data will be made available on request.

\section*{Acknowledgments}
We thank the Center for Information Technology of the University of Groningen for their support and for providing access to the Hábrók high-performance computing cluster.

 \bibliographystyle{elsarticle-num-names} 
 \bibliography{reference}





\end{document}